\documentclass[journal]{IEEEtran}
\usepackage{amsmath,amsfonts}
\usepackage{algorithmic}
\usepackage{algorithm}
\usepackage{array}
\usepackage[caption=false,font=normalsize,labelfont=sf,textfont=sf]{subfig}
\usepackage{textcomp}
\usepackage{stfloats}
\usepackage{url}
\usepackage{verbatim}
\usepackage{graphicx}
\usepackage{cite}
\usepackage{float}
\usepackage{subfig}
\usepackage{enumitem}
\usepackage{graphicx}
\usepackage{tikz}
\usepackage{comment}
\usepackage{amsmath,amssymb}
\usepackage{color}

\usepackage{booktabs}
\usepackage{bbding}
\usepackage{xspace}
\usepackage{interval}
\usepackage{bm}
\usepackage{enumitem}
\usepackage{mathtools}
\usepackage{footmisc}
\usepackage{tabularx}
\usepackage{multirow}

\usepackage{siunitx}
\usepackage[accsupp]{axessibility} 
\usepackage{makecell}

\setlist[enumerate,1]{leftmargin=*} 
\setlist[enumerate,0]{label=(\arabic*),widest=a}

\newcommand*{\eg}{\emph{e.g.}\@\xspace}
\newcommand*{\etc}{\emph{etc}\@\xspace}
\newcommand*{\ie}{\emph{i.e.}\@\xspace}
\newcommand*{\etal}{\emph{et al.}\@\xspace}

\definecolor{citecolor}{RGB}{0, 113, 188}
\usepackage[pagebackref=true,breaklinks=true,colorlinks,bookmarks=false, citecolor=citecolor]{hyperref}
\newcommand{\fakeparagraph}[1]{\vspace{3mm}\noindent\textbf{#1}}

\begin{document}

\title{Fighting Malicious Media Data: A Survey on Tampering Detection and Deepfake Detection}

\author{Junke Wang, Zhenxin Li, Chao Zhang, Jingjing Chen, Zuxuan Wu~\IEEEmembership{Member,~IEEE} \\ Larry S. Davis~\IEEEmembership{Fellow,~IEEE}, Yu-Gang Jiang~\IEEEmembership{Senior Member,~IEEE}

\IEEEcompsocitemizethanks{\IEEEcompsocthanksitem J. Wang, Z. Li, C. Zhang, J. Chen, Z. Wu and Y.G. Jiang are with School of Computer Science, Fudan University.
\IEEEcompsocthanksitem L. Davis is with University of Maryland.}
}

\maketitle

\begin{abstract}
Online media data, in the forms of images and videos, are becoming mainstream communication channels. However, recent advances in deep learning, particularly deep generative models, open the doors for producing perceptually convincing images and videos at a low cost, which not only poses a serious threat to the trustworthiness of digital information but also has severe societal implications. This motivates a growing interest of research in media tampering detection, \ie, using deep learning techniques to examine whether media data have been maliciously manipulated. Depending on the content of the targeted images, media forgery could be divided into image tampering and Deepfake techniques. The former typically moves or erases the visual elements in ordinary images, while the latter manipulates the expressions and even the identity of human faces. Accordingly, the means of defense include image tampering detection and Deepfake detection, which share a wide variety of properties. In this paper, we provide a comprehensive review of the current media tampering detection approaches, and discuss the challenges and trends in this field for future research.
\end{abstract}


\section{Introduction}
\label{sec:introduction}
\IEEEPARstart{R}{ecent} years have witnessed the rapid development of deep generative models~\cite{goodfellow2014generative, karras2019style, choi2018stargan}, which are able to generate perceptually convincing images and videos. While these techniques can benefit applications like AR/VR, creative designs, image editing, \etc, they are essentially a double-edged sword considering the potential impact on human society. For instance, the ubiquity of open-sourced tools based on deep generative models makes it extremely easy to create manipulated media data, \eg, inpainted images~\cite{lama-cleaner} or Deepfakes~\cite{fakeapp,deepfakes} that could be disseminated on the Internet for malicious purposes. This poses serious threats to the integrity of social applications like influencing president elections. 

The above challenges have stimulated a growing interest to automatically discriminate whether media data have been tampered or not with handcrafted features~\cite{barni2010identification,popescu2005exposing} and even deep learning techniques~\cite{zhou2018learning,wu2019mantra,hu2020span,li2020face,zhao2021multi}, with an aim to fight misinformation. In this paper, we classify the existing media forensics method into two categories: tampering detection and Deepfake detection. More specifically, tampering detection aims to detect whether generic objects have been added or removed in images and videos~\cite{ferrara2012image,zhou2018learning,bappy2017exploiting,hu2020span}, while Deepfake detection is more face-oriented that aims to  detect whether the expression or identity of human faces in images/videos has been manipulated~\cite{masi2020two, wu2020sstnet, zhao2021multi}. Current literature typically treats tampering detection and Deepfake detection as separate problems due to their differences in the overall process: (1) as Deepfake detection focuses on human faces, the input is typically a cropped face image, while tampering detection often takes in the entire image as inputs and aim to identify forged regions; (2) tampering detection requires to locate the manipulated areas, while most Deepfake detection methods only produce a binary Real/Fake prediction.

\begin{figure}[t]
  \centering
  \includegraphics[width=\linewidth]{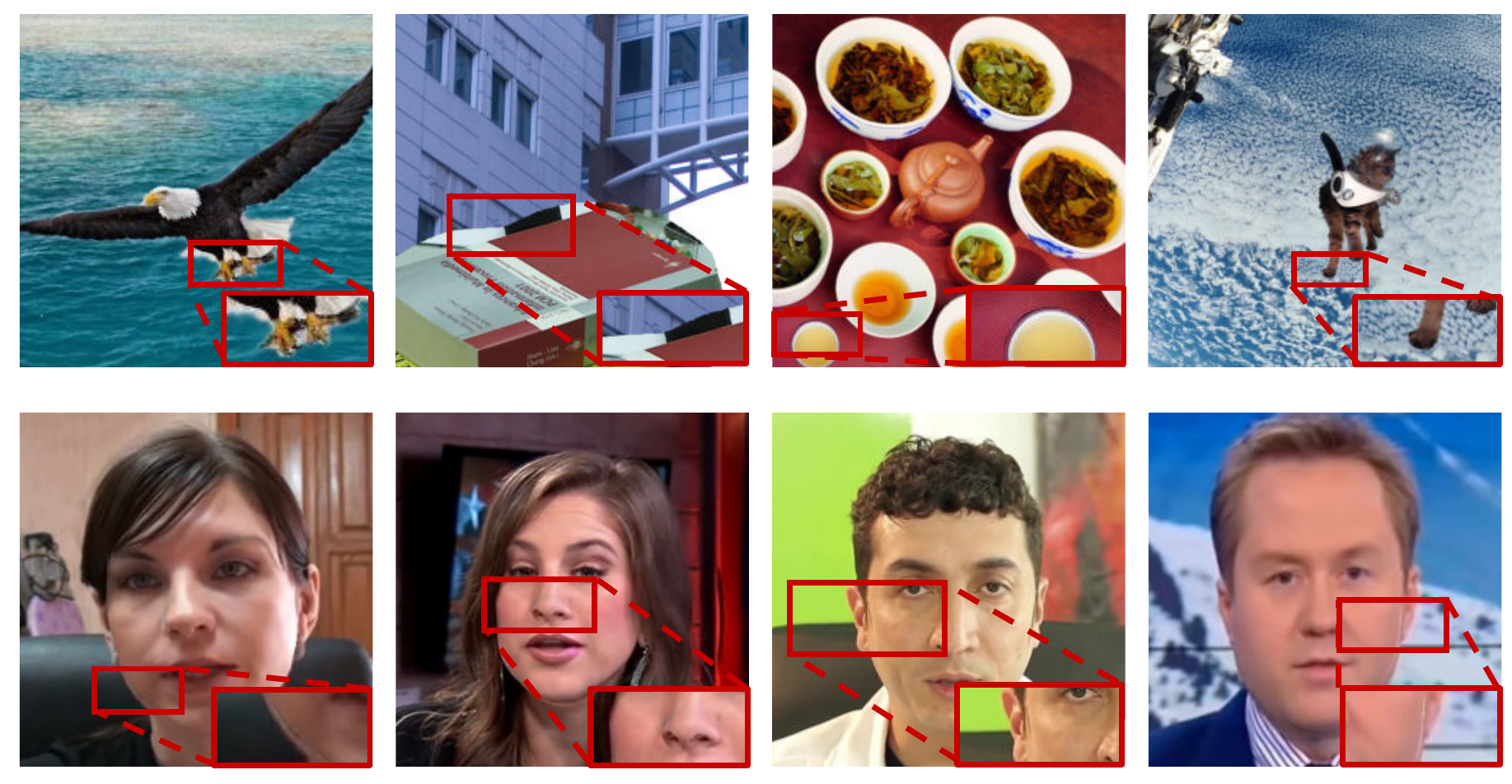}
  \caption{Visual artifacts of the manipulated images (top) and Deepfakes (bottom) in existing datasets.}
  \label{fig:artifacts}
\end{figure}

\begin{figure*}[t]
  \centering
  \includegraphics[width=\textwidth]{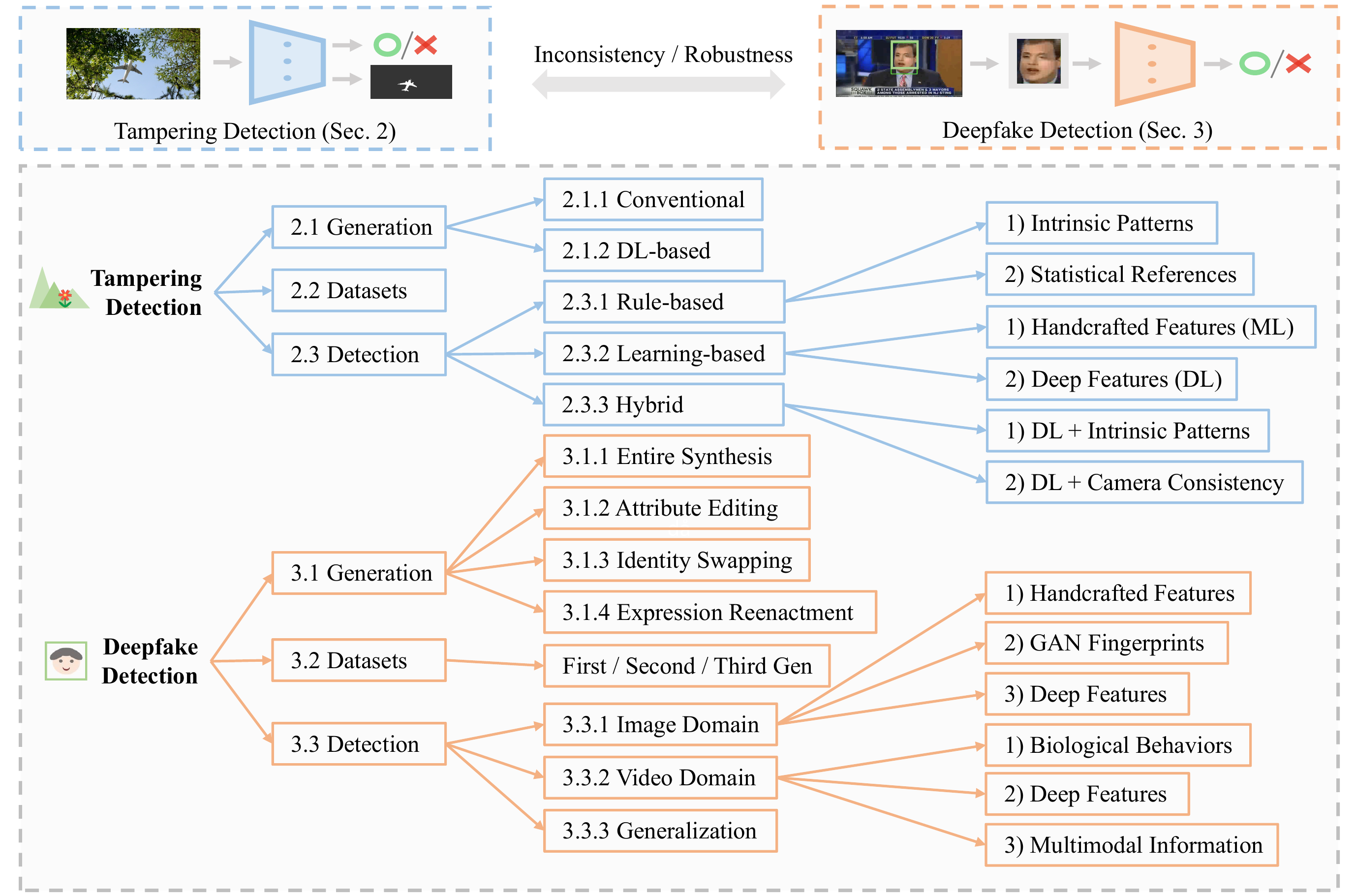}
  \caption{An overview of media forensics, which can be divided into tampering detection (TD) and Deepfake detection (DD). Typically, TD methods take a complete image as input and simultaneously identify the authenticity and locate the manipulated regions, while DD methods take a cropped face as input and produce a binary Real/Fake prediction. }
  \label{fig:outline}
  \vspace{1em}
\end{figure*}

But essentially, both tampering detection and Deepfake detection are discriminative tasks that attempt to discover forgery traces through careful examination of visual contents. As a result, they both highly rely on (1) inconsistency modeling: the imaging principle of the camera determines that the pixels of pristine images follow a certain statistical distribution, which generative models typically struggle to reconstruct~\cite{yu2019attributing}. Therefore, media tampering including image manipulation and Deepfake, followed by blending techniques that mix altered objects/faces with background images, will lead to intrinsic inconsistency, between the manipulated and authentic regions. This inspires forensics approaches to carefully examine the visual artifacts (see Figure~\ref{fig:artifacts}) in suspicious images to discover both global~\cite{wang2022objectformer} and local~\cite{li2020face,zhao2021learning,chen2021local} inconsistent information and (2) robust features: media data inevitably go through various post-processing during its creation and dissemination. In addition, malicious users will deliberately impose perturbations on the tampered images to fool the detection tools. This could erase the manipulation traces and significantly increase the difficulty for forensics, requiring the defending approaches to capture more robust and meaningful forgery evidence.

The common nature shared by tampering detection and Deepfake detection motivates us to present a unique and comprehensive survey of both fields in this literature to facilitate future research. While there are a few concurrent surveys, they typically put a primary focus on a single aspect of tampering detection~\cite{verdoliva2020media, birajdar2013digital} or Deepfake detection~\cite{mirsky2021creation,nguyen2022deep,malik2022deepfake}. Our work differs in that we believe tampering detection and Deepfake detection are similar in spirit and thus we summarize the newest advances of both fields systematically, and hope that future research can learn from the best of both worlds. 

The remainder of this survey is organized as follows: in Sec.~\ref{sec:tampering_detection} and Sec.~\ref{sec:deepfake_detection}, we separately introduce the progress in the field of tampering detection and Deepfake detection, including manipulation techniques, public datasets for evaluation, and various detection techniques. Sec.~\ref{sec:challenges} presents the remaining challenges in the field of tampering detection, and outlines the future research directions. Finally, we briefly conclude this paper in the following Sec.~\ref{sec:conclusion}. The outline of this survey is illustrated in Figure~\ref{fig:outline}.

\section{Tampering Detection}
\label{sec:tampering_detection}
With the rise of data-driven techniques and the vast amount of media content within easy reach, media tampering methods~\cite{wang2022manitrans, lee2022sound, kim2022diffusionclip} allow users to edit or create realistic images automatically, and as a result, there grows an urgent need for effective and trustworthy detection methods~\cite{wang2022objectformer, bammey2022forgery, wu2022robust} that can distinguish fake images from real ones to preserve the credibility of media content. In this section, we introduce the advancements made in the generation (Sec.~\ref{subsec:tampering_generation}), the datasets (Sec.~\ref{subsec:tampering_datasets}) and the detection (Sec.~\ref{subsec:tampering_detection}) of tampered images. For tampering detection, our main focus is on the images whose subjects are generic and not constrained to a particular type, while for the following section Deepfake detection (Sec.~\ref{sec:deepfake_detection}), we dive into the detection of manipulated human faces.

\subsection{Tampering Generation}
\label{subsec:tampering_generation}
\begin{figure}[t]
  \centering
  \includegraphics[width=\linewidth]{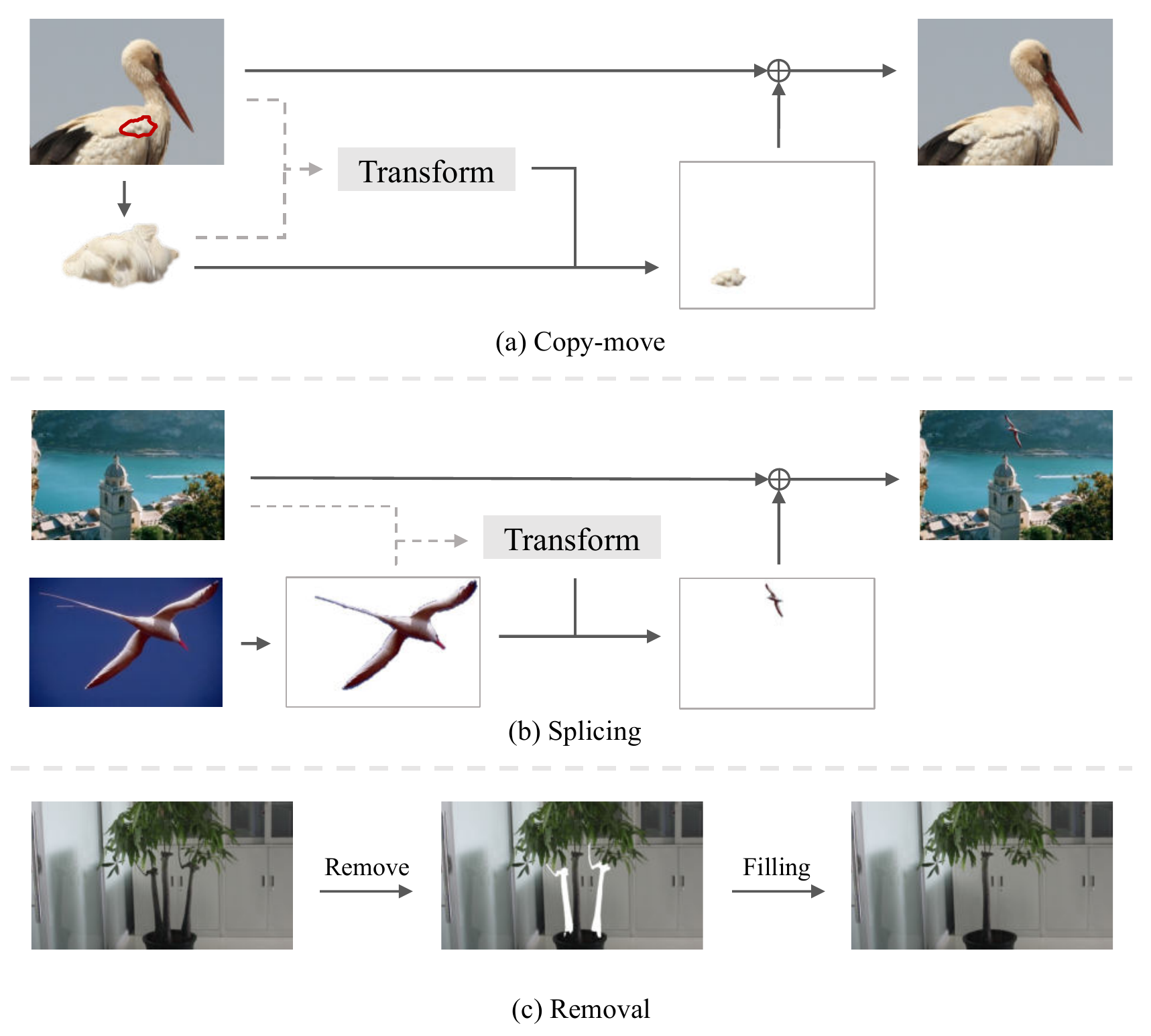}
  \caption{Illustration of the process to manipulate an image through (a) copy-move, (b) splicing, and (c) removal.}
  \label{fig:tampering_generation}
\end{figure}

From the earliest automatic inpainting methods that remove unwanted elements from images~\cite{bertalmio2000image}, much research in generating images with authenticity based on provided source objects (\eg, splicing, editing, removal of subjects) has been carried out. Early approaches~\cite{criminisi2004region, criminisi2003object, barnes2009patchmatch} focus on the texture and structure information presented in images to reconstruct target image patches or regions. Recently, deep learning-based generative models, \eg, GANs~\cite{goodfellow2014generative} and diffusion models~\cite{ho2020denoising, dhariwal2021diffusion}, are also proposed to achieve realistic synthesis without devoting too much effort to the analysis of the intrinsic distributional information. This also opens up new opportunities for more user-interactive manipulations~\cite{kim2022diffusionclip, wang2022manitrans} since researchers can focus more on how to embed user interactions into the model rather than merely improve the synthesis quality. Based on the algorithms adopted, tampering generation can be categorized into conventional methods (Sec.~\ref{subsubsec:tampering_conventional_generation}) and deep learning-based methods (Sec.~\ref{subsubsec:tampering_dl_generation}). We illustrate the pipeline of image tampering approaches in Figure~\ref{fig:tampering_generation}.

\subsubsection{Conventional Methods}
\label{subsubsec:tampering_conventional_generation}
Conventional tampering methods conduct forgeries based on the information within the same image, which can be cheaply calculated and directly applied to the target region. Even though these methods do not require training on large amounts of data, they are still sometimes time-consuming due to the computational complexity of the algorithm, \eg, the cost of matching similar patches~\cite{barnes2009patchmatch}, which may significantly hinder user interactions.

Specifically, \cite{criminisi2003object, criminisi2004region} can handle the removal of large objects in images through exemplar-based synthesis, which iteratively select a template region along the target contour to inpaint given the best matching patch with similar textures and structures. After each iteration, the target region is shrunk as the template region is filled with new content and its contour is propelled inward. Furthermore, \cite{barnes2009patchmatch} proposes a randomized algorithm to optimize the process of patch-matching, which first assigns some initial guesses of similar patches and then refines the best-fitting patch through random searches in its concentric neighborhoods.

\subsubsection{Deep Learning-Based Methods}
\label{subsubsec:tampering_dl_generation}
Compared with conventional methods, deep learning-based tampering methods, typically deep generative models like GANs~\cite{goodfellow2014generative}, are able to synthesize realistic images by training a generator and a discriminator in an adversarial manner. Conditional GANs~\cite{mirza2014conditional}, on the other hand, take additional driving signals as inputs to guide the creation of images. With these approaches, manipulating images in a data-driven manner becomes feasible.
Here, we divide these deep learning-based tampering methods into three categories in terms of their manipulation objective, \ie, splicing, object editing, and object removal.

\fakeparagraph{Splicing} copies regions from different source images and pastes them to the target image. Built upon the GAN-based architecture, several approaches~\cite{lin2018st, zhan2019spatial, li2021image} adopt the Spatial Transformer Network in their architecture to warp the source object into the background geometrically, and~\cite{zhan2019spatial, li2021image} further model the appearance patterns of sources to obtain appearance-preserving results.
To better maintain the characteristics of sources, \cite{azadi2020compositional} first combines two images and then separates them using a decomposition network, so the reconstructed objects serve as an additional supervisory signal.
\cite{hong2018learning, tan2019image} both utilize semantic maps and user input to combine source images. Specifically, \cite{tan2019image} estimates depth information and allows users to decide the occlusion relationships between different objects.

\fakeparagraph{Object Editing} aims to manipulate the attributes of objects according to the instructions of users. We particularly focus on multi-modal object editing (\ie text-based~\cite{li2020manigan, wang2022manitrans, kim2022diffusionclip}, sound-based ~\cite{lee2022sound}) in this part. 
A text-driven generative model~\cite{li2020manigan, wang2022manitrans, kim2022diffusionclip} typically contains an image encoder and a text encoder, from which the latent representations of the source image and the driving information are respectively derived. To fuse the representations, \cite{li2020manigan} proposes a text-image affine combination module, which proves to be more effective than simply concatenating features. To further generate entity-level believable results, \cite{wang2022manitrans} adopts the transformer architecture~\cite{vaswani2017attention} with tokenized text and image as input to model the relationships between 
different modalities, while \cite{kim2022diffusionclip} combines a diffusion model~\cite{ho2020denoising, dhariwal2021diffusion} with CLIP and allows zero-shot image manipulation between unseen domains.
Furthermore,~\cite{lee2022sound} extends the CLIP-based manipulation methods~\cite{patashnik2021styleclip, kim2022diffusionclip} with an additional audio branch and a shared multi-modal latent space to embed triplet pairs (\ie, image, text, and sound). The derived latent code is then given as input to a StyleGAN2~\cite{karras2020analyzing} generator alongside encoded audio features to generate the final image.
Besides the aforementioned cross-modal editing methods, \cite{dhamo2020semantic} generates a semantic scene graph, whose nodes and edges respectively represent the objects present in the image and their relations, to model the target image in a semantically manipulable manner.

\fakeparagraph{Object Removal}, introduced in Sec.~\ref{subsubsec:tampering_conventional_generation}, erases regions from
images and inpaints missing regions with visually plausible
contents. Observed by \cite{shetty2018adversarial}, relying on the generator itself to perform object removal tends to lead to a completely re-synthesized image, while we only desire to manipulate one region. To address this issue, \cite{shetty2018adversarial} proposes a two-stage architecture composed of a mask generator and an inpainting network, which are jointly trained to achieve satisfactory removal results. It is worth noting that removing a node in the scene graphs produced by~\cite{dhamo2020semantic} can also perform object removal in an image.

\begin{table}[t]
\renewcommand\arraystretch{1.1}
\centering
  \caption{Basic information of existing image tampering detection datasets.}
  \label{tab:tampering_dataset}
    \setlength{\tabcolsep}{0.pt} 
    \begin{tabular*}{\linewidth}{@{\extracolsep{\fill}}lccccccc@{}}
  \toprule
    \multirow{2}{*}{\textbf{Dataset}} && \multicolumn{2}{c}{\textbf{\makecell[c]{Images}}} && \multicolumn{3}{c}{\textbf{\makecell[c]{Manipulations}}} \\
    ~ && \makecell[c]{Real} & \makecell[c]{Forged} && \makecell[c]{splicing} & \makecell[c]{copy-move} & \makecell[c]{removal} \\
    \cmidrule{1-1} \cmidrule{3-4} \cmidrule{6-8}
    Columbia Gray~\cite{ng2004data} && 933 & 912 && \checkmark & ~ & ~ \\
    Columbia Color~\cite{hsu2006detecting} && 183 & 180 && \checkmark & ~ & ~ \\
    MICC-F8multi~\cite{amerini2011sift} && - & 8 && ~ & \checkmark & ~ \\
    MICC-F220~\cite{amerini2011sift} && 110 & 110 && ~ & \checkmark & ~ \\
    MICC-F600~\cite{amerini2011sift} && 440 & 160 && ~ & \checkmark & ~ \\
    MICC-F2000~\cite{amerini2011sift} && 1300 & 700 && ~ & \checkmark & ~ \\
    VIPP Synth.~\cite{bianchi2012image} && 4800 & 4800 && \checkmark & ~ & ~ \\
    VIPP Real.~\cite{bianchi2012image} && 69 & 69 && \checkmark & ~ & ~ \\
    CoMoFod~\cite{tralic2013comofod} && 260 & 260 && ~ & \checkmark & ~ \\
    CASIA V1.0~\cite{dong2013casia} && 800 & 921 && \checkmark & \checkmark & ~ \\
    CASIA V2.0~\cite{dong2013casia} && 7200 & 5123 && \checkmark & \checkmark & ~ \\
    Wild Web~\cite{zampoglou2015detecting} && 90 & 9657 && \checkmark & ~ & ~ \\
    NC2016~\cite{guan2019mfc} && 560 & 564 && \checkmark & \checkmark & \checkmark \\
    NC2017~\cite{guan2019mfc} && 2667 & 1410 && \checkmark & \checkmark & \checkmark \\
    MFC2018~\cite{guan2019mfc} && 14156 & 3265 && \checkmark & \checkmark & \checkmark\\
    MFC2019~\cite{guan2019mfc} && 10279 & 5750 && \checkmark & \checkmark & \checkmark \\
    PS-Battles~\cite{heller2018ps} && 11142 & 102028 && \checkmark & \checkmark & \checkmark\\
    DEFACTO~\cite{mahfoudi2019defacto} && - & 229000 && \checkmark & \checkmark & \checkmark\\
    IMD2020~\cite{Novozamsky_2020_WACV} && 35000 & 35000 && \checkmark & \checkmark & \checkmark \\
    \bottomrule
    \end{tabular*}
\end{table}

\subsection{Tampering Datasets}
\label{subsec:tampering_datasets}
In order to facilitate the development of tampering detection, various datasets are proposed to benchmark detection approaches. Table~\ref{tab:tampering_dataset} presents the commonly used tampering datasets as well as their corresponding manipulation approaches.

\begin{figure}[t]
  \centering
  \includegraphics[width=\linewidth]{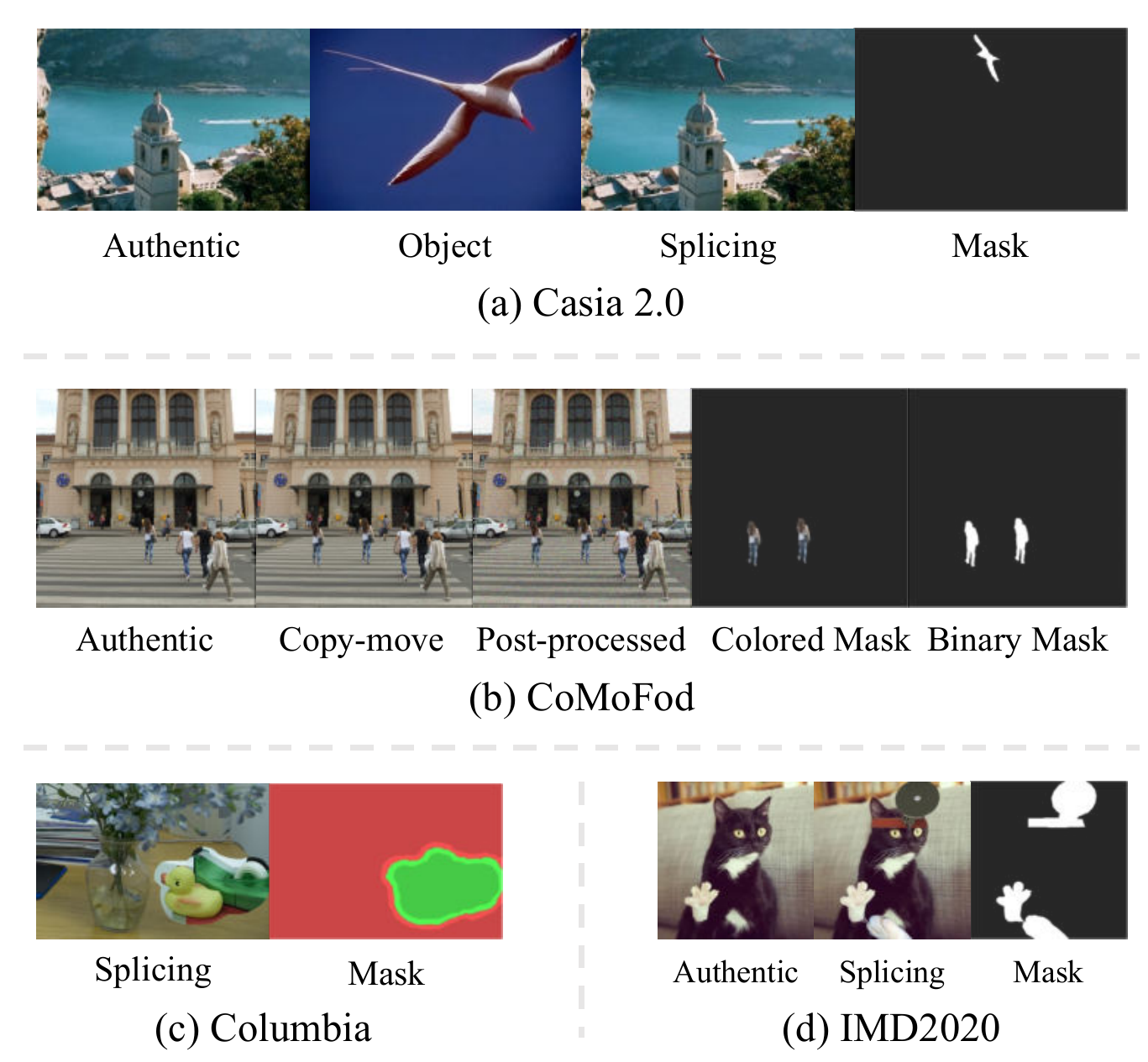}
  \caption{Examples from the existing tampering datasets. We display both tampered images and masks.}
  \label{fig:tampering_datasets}
\end{figure}

Since the release of the earliest \textbf{Columbia}~\cite{ng2004data, hsu2006detecting} comprising splicing forgeries, numerous datasets focusing on other manipulations, such as copy-move~\cite{amerini2011sift, christlein2012evaluation, bianchi2012image, tralic2013comofod} have been developed. \textbf{CASIA v1.0 and  CASIA v2.0} ~\cite{dong2013casia} are among the first to include two kinds of manipulations in one dataset. Even though CASIA v2.0~\cite{dong2013casia} is already much larger (\ie, 7200 real images and 5123 forged images) than previously released datasets, it was not scalable enough because the forged images were manually crafted by a group of image makers using Adobe Photoshop. To produce a large number of tampered images and add more comprehensiveness to the data, the \textbf{Wild Web}~\cite{zampoglou2015detecting} collects forged images from the Internet and its scale greatly surpasses the aforementioned datasets, while the \textbf{PS-Battles}~\cite{heller2018ps} similarly gathers an extensive number of images edited by Adobe Photoshop from Reddit communities. In addition, the \textbf{MFC Datasets}~\cite{guan2019mfc} released by NIST involve a series of comprehensive datasets (\ie, NC2016, NC2017, MFC2018, MFC2019), which sets a notable benchmark for the evaluation of media tampering detection. \textbf{DEFACTO}~\cite{mahfoudi2019defacto} and \textbf{IMD2020}~\cite{Novozamsky_2020_WACV} are the other two recently published large-scale datasets, and DEFACTO obtains its sources from the Microsoft COCO dataset~\cite{lin2014microsoft}, while IMD2020 derives its samples from manually selected online images that are free of distinct manipulation traces. The latest \textbf{ImageForensicsOSN} dataset~\cite{wu2022robust} leverages tampered images from multiple sources~\cite{guan2019mfc, de2013exposing, hsu2006detecting, dong2013casia} and produces OSN-transmitted (\ie, transmitted via an online social network, which induces noises in images) versions of them to facilitate detection methods that are robust against OSN-transmissions. We demonstrate several samples from the above tampering datasets~\cite{hsu2006detecting, tralic2013comofod,dong2013casia, Novozamsky_2020_WACV} in Figure~\ref{fig:tampering_datasets}.

\subsection{Tampering Detection}
\label{subsec:tampering_detection}
In this section, we elaborate on media tampering detection, which is a research field of growing importance since deep generative models described in Sec.~\ref{subsubsec:tampering_dl_generation} exhibit effectiveness and handiness. Nevertheless, the emergence of deep learning benefits not only manipulation tools but also forensic techniques by sparing forgery analysts from handcrafting features or rules for forensic purposes. Though the combination of deep learning (DL) and tampering detection marks a notable milestone in the field, DL-based detectors lack interpretabilities, yielding questionable results that 
can be difficult to accept on serious occasions such as a court.

In an effort to frame the problem of tampering detection systematically, we take both current research trends and classic forensic techniques in early works into account and thus formulate three major categories for tampering detection. The first and second class, \ie Rule-based (Sec.~\ref{subsubsec:tampering_rule_detection}) and Learning-based (Sec.~\ref{subsubsec:tampering_learning_detection}) methods, are categorized based on whether machine learning (ML) or deep learning (DL) algorithms are utilized to detect tampering clues.
On the other hand, we find that some methods gracefully combine the rule-based methods devised by forgery analysts with neural networks, which make them different from methods falling into the previous two categories. As a result, we come up with the third category, \ie Hybrid Detection Methods (Sec.~\ref{subsubsec:tampering_hybrid_detection}), alongside the previous two to present readers the whole picture of tampering detection.

\subsubsection{Rule-based Detection Methods}
\label{subsubsec:tampering_rule_detection}
For rule-based detection methods, we classify them into intrinsic patterns analysis and methods with statistical references. Considering the differences in the inference process of these methods, the first category requires no prior knowledge of image samples and carries out tampering detection on them straightly, while methods from the second category estimate whether an image is tampered with based on statistics of pristine samples.

\fakeparagraph{Intrinsic Patterns Analysis} is a kind of detection techniques that explores the underlying patterns within image samples. By identifying the abnormalities induced by image tampering, it carries out tampering detection exclusively on the image samples without the help of additional prior knowledge. The intrinsic patterns of a naturally produced image result from the acquisition chain and the encoding process, and the consistency in these patterns can be disrupted after the image is manipulated.

Different color components tend to focus on different lateral offsets on the sensor when an image is captured by the camera's optical lens due to their varied wavelengths. This phenomenon, named lateral chromatic aberration (LCA), can be exploited for tampering detection~\cite{gloe2010efficient, mayer2018accurate} by identifying the inconsistent LCA displacement in manipulated regions. Nevertheless, LCA can be eliminated by image post-processing, which may invalidate these methods. 

After the light goes through the lens, it becomes post-processed by a color filter array, which filters light of different wavelengths and separates color information for the image sensor. This process is generally followed by a procedure known as demosaicing, where color interpolation is performed to fill the missing pixel values for the final image, leaving detectable demosaicing traces on the image as a fingerprint. The lack of such fingerprints in an image region usually suggests the occurrence of image tampering. Following~\cite{popescu2005exposingarray}, studies 
made by~\cite{cao2009accurate, ho2010inter} seek to model the periodic pattern produced by demosaicing. Apart from leaving demosaicing traces on the image, the image sensor, as well as the rest of the acquisition chain, tends to impose unwanted noise on the image. Despite its randomness, the noise can similarly be modeled as a periodic pattern. \cite{liu2014digital, cozzolino2015splicebuster} reveal splicing and copy-move forgeries by detecting irregular noise levels, while~\cite{swaminathan2008digital} applies noise residuals to the estimation of a camera-imposed fingerprint.

Another crucial step in acquiring the image content suitable for transmission is compression, especially JPEG compression, which happens both at the device level and the software level. When a compressed image is manipulated and then saved by a photo-editing software, \eg, Adobe Photoshop, it is compressed again, leaving detectable double compression artifacts on the image~\cite{popescu2004statistical}. However, double compression artifacts alone do not suffice to indicate the presence of image tampering since the user may only save the image without editing it. To address this issue, ~\cite{barni2010identification} points out that the tampered regions are very likely to be compressed again under misaligned grids, leaving detectable traces for tampering detection, while~\cite{chen2011detecting} performs periodicity analysis on double-compressed images in both spatial and transform domains, which proves robust enough against different alignment situations. Besides,~\cite{agarwal2017photo, pasquini2017statistical} look into the histograms of DCT (Discrete Cosine Transform) coefficients in JPEG compression to discover statistical inconsistencies, while~\cite{farid2009exposing} inventively compresses the image further with different JPEG qualities to expose the ``JPEG ghost''. Aside from the previously mentioned methods,~\cite{iakovidou2018content} analyzes the patterns of blocking artifacts produced by JPEG compression, while~\cite{lorch2019image} investigates the artifacts originating from chroma sub-sampling in JPEG compression.

Eventually, to create a realistic tampered image, editors need to make the tampered region blend in with its background. Common photo editing approaches such as blurring, shape and color transformations are often adopted during handcrafted image tampering, which may lead to detectable manipulation traces. For the splicing detection techniques~\cite{dong2008run, bahrami2015blurred}, whose test images are the combination of multiple sources rather than single-sourced, the inconsistent pixel correlations~\cite{dong2008run} and blur types~\cite{bahrami2015blurred} are explored. As for copy-move detection, many methods seek to find the repetitive occurrence of the same image component. The earliest method proposed in~\cite{fridrich2003detection} detects copy-move manipulations by matching and comparing small image segments. Later, \cite{bayram2009efficient, ardizzone2010copy, amerini2011sift, christlein2012evaluation, amerini2013copy, ryu2013rotation, li2014segmentation, silva2015going, ardizzone2015copy} all perform copy-move forensics by either utilizing image keypoints~\cite{amerini2011sift, amerini2013copy, li2014segmentation, ardizzone2015copy, silva2015going}, or image block features~\cite{bayram2009efficient,ardizzone2010copy, ryu2013rotation}. By using hierarchical feature point matching, \cite{li2018fast} improves keypoint-based methods in terms of their insufficient keypoint sampling and their inabilities to cope with regions with varied sizes and textures.

\fakeparagraph{Methods with Statistical References} refer to the forensic methods that require prior knowledge related to the image acquisition chain as detection guidance. Here, we specifically discuss the methods using the PRNU (\ie, photo-response non-uniformity) noise, which can be regarded as a sensor-specific fingerprint. The major difference between this genre and the aforementioned intrinsic patterns analysis consists in the fact that this genre first approximates a pattern of an imaging sensor given a statistical sample and then compares the pattern of test image with it to derive a detection result~\cite{lukavs2006detecting, chen2007imaging}.
It is worth noting that the test image is preprocessed with a denoising filter to remove low-frequency components of PRNU and high-level scene content to speed up PRNU estimation~\cite{lukavs2006detecting}. However, PRNU signals are generally weak and thus hinder the detection of small-sized forgeries, so~\cite{chierchia2014guided} designs an adaptive filtering technique to enhance the resolution of PRNU-based methods. Other methods proposed in~\cite{chierchia2014bayesian, chakraborty2017prnu} are further integrated with machine learning algorithms.

\subsubsection{Learning-based Detection Methods}
\label{subsubsec:tampering_learning_detection}
In this section, we further expand on tampering detection by discussing learning-based forensic tools, among which machine learning-based methods will first be introduced, followed by today's trending topic, \ie, deep learning-based tampering detectors.

\fakeparagraph{Machine Learning Methods with Handcrafted Features} require the researchers to investigate the manipulation traces left on a forged image, which are further analyzed by an ML algorithm to perform a binary Pristine/Fake classification. Frequency domain statistics are widely explored to capture the subtle forgery artifacts by DCT transformation~\cite{he2006detecting}, DWT transformation\cite{lyu2005realistic,chen2007image}, and even their combinations~\cite{he2012digital}. In addition, ~\cite{lin2005detecting, hsu2006detecting, hsu2010camera} model the camera characteristics consistency, and~\cite{wang2009effective, ke2014detecting} investigate features from the color space of an image. With these handcrafted features available, ~\cite{bayram2006image, he2012digital, cozzolino2014imageres, cozzolino2014imagefusion, ferreira2016behavior} further adopt different fusion strategies to produce the final predictions, which achieves more competitive results than operating on a single type of features. Specifically, \cite{bayram2006image, he2012digital} extract multiple types of features and select them accordingly to construct the best feature set, while~\cite{cozzolino2014imageres, cozzolino2014imagefusion} leverage decisions made by various approaches (\eg, the ML algorithm, the patch-matching algorithm, and the PRNU-based method), and~\cite{ferreira2016behavior} combines classifiers with the BKS (Behavior-Knowledge Space) method~\cite{huang1995method}.

\fakeparagraph{Deep Learning Methods} have dominated the field of tampering detection demonstrating superior performance since their emergence. An extensive number of studies are carried out to explore different network structures or training methods to achieve better results. A line of work~\cite{wu2017deep,salloum2018image,bi2019rru} focuses on a specific type of image tampering (\eg splicing and copy-move). For the splicing detection, ~\cite{wu2017deep} performs deep matching between the features from a query image and a potential donor image (\ie, a part of the donor's region may be spliced with the query image) extracted by two weight-sharing CNNs. Besides, ~\cite{salloum2018image} proposes a multi-task CNN to simultaneously predict an edge mask and a commonly used splicing mask to develop more sharp boundaries, and \cite{cun2018image} leverages global and local features derived from the entire RGB image and image patches. In addition,~\cite{bi2019rru} renovates the structure of the U-Net~\cite{ronnebergerconvolutional} by adding a residual feedback connection to a ResNet layer~\cite{he2016deep} for feature enhancement. 
For the copy-move detection,~\cite{wu2018image} proposes a Self-Correlation module to calculate the similarity between different pixels in a feature map, which is further utilized to localize the repeated regions. Furthermore,~\cite{wu2018busternet} proposes a two-branch network where one branch localizes only the manipulated area while the other one follows~\cite{wu2018image} to predict cloned regions. ~\cite{zhong2019end} designs a network with dense feature connections and a key-point-based Feature Correlation, which outperforms~\cite{wu2018busternet} on unseen manipulated objects during the training stage. Building upon the attention mechanism,~\cite{islam2020doa} designs a Dual-Order Attention module that first computes two attention maps based on image self-correlation, with which attentive features are calculated by element-wise product and matrix multiplication. Given these robust features, the localization network is then trained in an adversarial manner with a PatchGAN discriminator~\cite{isola2017image}, which improves the localization ability of the generator and the image encoder. Also involving generative models, \cite{cozzolino2016single} uses an Autoencoder to learn the data distribution of pristine images by performing image reconstruction and discriminative labeling on encoded features to locate the anomalies without training on forged images.

Instead of focusing on one particular tampering method, universal forensic tools~\cite{bayar2016deep} are agnostic to the tampering method, as well as what pre- and post-processing operations are used and which camera the image is captured by. Without the constraints of certain assumptions, such methods are able to detect various forgeries in a single model. An early method proposed in~\cite{zhang2016image} relies only on MLPs and Stacked Autoencoders~\cite{vincent2010stacked} to detect anomalous patches, while methods in~\cite{bayar2016deep, bayar2018constrained, zhang2018boundary} base their analysis on CNNs and make adaptions to them for better detection results. In particular, \cite{bayar2016deep, bayar2018constrained} redesign the convolutional layer to suppress image content and highlight manipulation features, while~\cite{zhang2018boundary} designs a CNN consisting of only 2 convolutional layers and no pooling layers to better distinguish the boundaries of forged regions.
Moreover, to overcome the limited receptive field of a CNN, an LSTM is appended to the convolutional layers in~\cite{bappy2017exploiting, bappy2019hybrid} to further model the interdependencies of different blocks in 2D feature maps, and thus fosters the learning of boundary transformation between neighboring blocks. Concerning how image quality affects tampering detection,~\cite{marra2020full} points out that resizing the image to fit it into the GPU memory ruins high-frequency details that are precious for media forensics, while studies made by~\cite{rao2021self, wu2022robust} bring forward methods making detectors robust against JPEG-compressed images~\cite{rao2021self} and images transmitted via OSN (Online Social Network)~\cite{wu2022robust}.~\cite{wu2019mantra} considers numerous manipulation types (up to 385) and trains the network to learn manipulation traces via self-supervised manipulation classification. It should be clarified that compared with the several tampering methods included in our survey (Sec.~\ref{fig:tampering_generation}), the 385 manipulation types in~\cite{wu2019mantra} are categorized according to what editing operation is used and its corresponding parameters.

\begin{figure}[t]
  \centering
  \includegraphics[width=\linewidth]{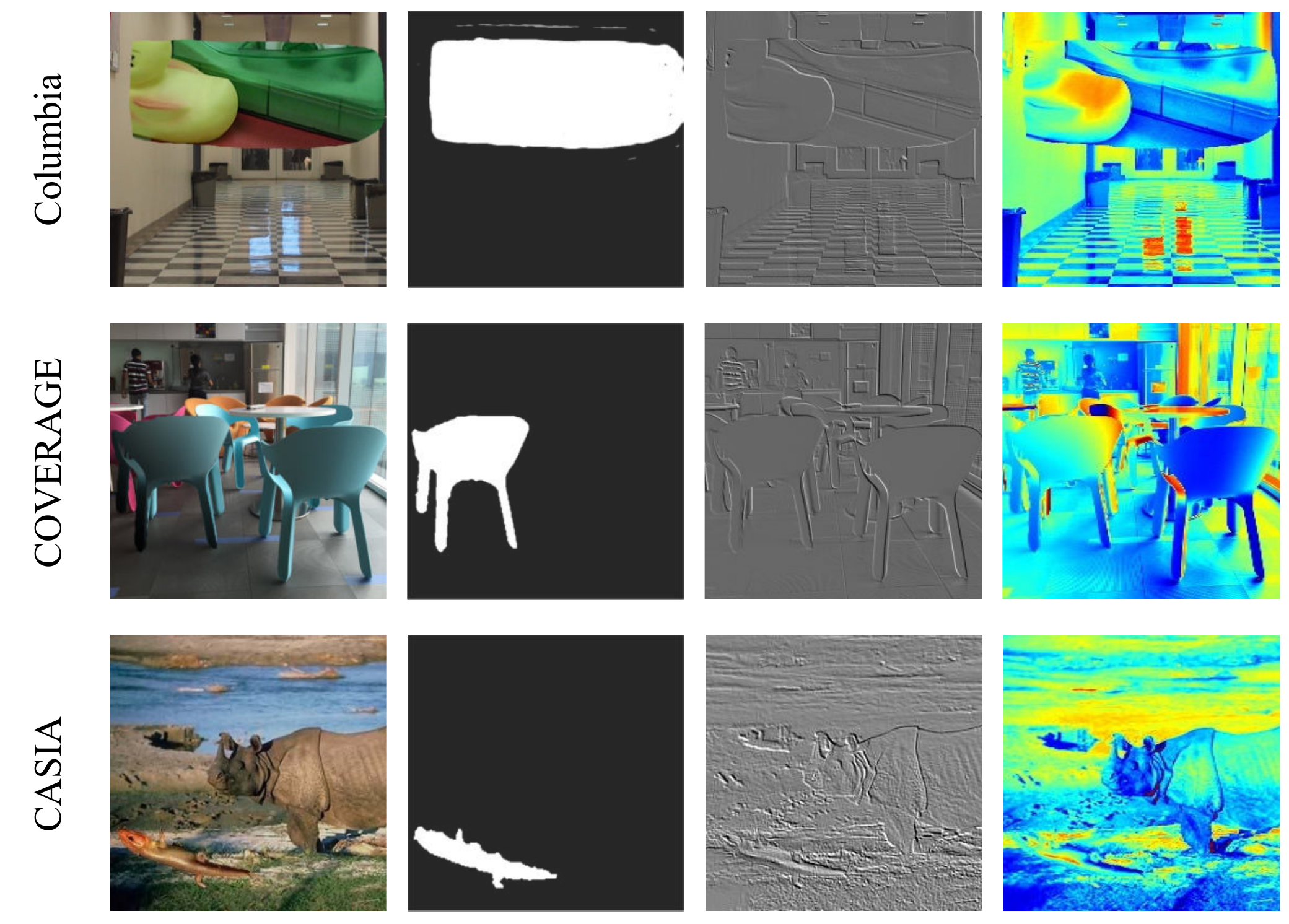}
  \caption{Visualization of several handcrafted features for image tampering detection and localization. From left to right, we show the manipulated images, masks, noise, and high-frequency components, respectively.}
  \label{fig:noise_and_frequency}
\end{figure}

By integrating features of different scales and modalities, universal forensic methods achieve more appreciable performances. In~\cite{liu2018image, hu2020span, hao2021transforensics, chen2021image, liu2022pscc}, the detectors are constructed to learn the relationships between image patches at different scales, which are produced by different layers of the CNN backbone~\cite{hao2021transforensics, chen2021image}, extracted from patches with different sizes~\cite{liu2018image}, or derived from a pyramid structure~\cite{hu2020span, liu2022pscc}. Rather than predict a single manipulation mask with fused multi-scale features, ~\cite{liu2022pscc} first derives a mask from the smallest scale and then progressively refines it to larger scales.
Meanwhile, multi-modal detectors~\cite{zhou2018learning, mazaheri2019skip, bappy2019hybrid, wang2022objectformer} tackle forgeries by combining features from multiple domains, \eg, RGB domain, noise domain~\cite{zhou2018learning}, and frequency domain~\cite{mazaheri2019skip, bappy2019hybrid, wang2022objectformer}, as is visualized in Figure~\ref{fig:noise_and_frequency}. In~\cite{mazaheri2019skip, bappy2019hybrid} frequency features extracted by LSTM are fused with spatial features produced by convolutional layers, while~\cite{wang2022objectformer} combines RGB features with high-frequency features as multimodal patch embeddings, which are further refined by learnable object prototypes, and as a result, patch-level and object-level consistencies are captured in such cross-modal interactions. 

\subsubsection{Hybrid Detection Methods}
\label{subsubsec:tampering_hybrid_detection}
Even though DL-based forensic techniques can work well without engineering suitable rules for tampering detection, several studies look in a different direction, integrating deep learning with a priori knowledge and handcrafted rules of image tampering that is considered in Rule-based Detection Methods (Sec.~\ref{subsubsec:tampering_rule_detection}). Guided by extra knowledge and rules, hybrid detectors can not only pay more attention to tampering-related details but also are equipped with better interpretabilities since they usually give more specific predictions, which can be used for further analysis.

\fakeparagraph{DL-based Methods examining Intrinsic Patterns} add conventional elements to the data-driven scenario in deep learning. They generally operate on handcrafted features as inputs rather than raw pixels and rely on the strong capability of deep neural networks to learn the decision boundary from these features. A good case in point is a median filtering forensic method proposed in~\cite{chen2015median}, as it employs a filter layer to extract median filtering residuals from RGB images before the CNN model.

As is mentioned in Sec.~\ref{subsubsec:tampering_rule_detection}, double JPEG compression leaves distinct artifacts on the manipulated image and can be exploited for tampering detection. \cite{wang2016double, amerini2017localization, barni2017aligned, park2018double} uses deep learning methods to expose double compression artifacts by feeding the CNN with DCT coefficient histograms. \cite{amerini2017localization} designs two encoders respectively for RGB patches and DCT coefficient histograms and concatenates these cross-modal features before proceeding to the fully connected network, which demonstrates superior performance to single-modal network under various JPEG quality factors. Furthermore, to prepare the detection model for real-world applications, \cite{park2018double} generates a dataset of mixed JPEG qualities and handles these varied qualities in one model, in which the vectorized quantization table is concatenated with DCT histogram features for enrichment.

The second type operates on image pixels but is supervised with unique signals to expose certain artifacts. To detect demosaicing artifacts, an innovative training method named Positional Learning is proposed in~\cite{bammey2020adaptive, bammey2022forgery}, which consists in training a CNN to predict the modulo-2 position of image pixels. This self-supervised training method enables the CNN to be aware of the underlying periodic mosaic pattern, so a well-trained CNN's misprediction indicates inconsistencies in the pattern and therefore reveals forgeries. In specific, ~\cite{bammey2020adaptive} trains the network on a database of pristine images, while~\cite{bammey2022forgery} further extends~\cite{bammey2020adaptive}'s work and shows it is feasible to fine-tune the network on a single test image, whether pristine or forged, to adapt the network's weights to its domain. Even though the manipulated regions of a forged test image can mislead the network, these regions are generally small compared with the pristine regions, and as a result, the network can still benefit from the retraining process, which is also verified in~\cite{bammey2022forgery}'s ablation studies.

\fakeparagraph{DL-based Methods based on Camera Consistency} refer to the neural networks looking at disrupted camera consistency incurred by image tampering. An analogy can be drawn between these methods and those with statistical references in the rule-based section as both types first require training (or statistical estimation) to construct a camera-aware model and then make decisions based on learned camera information.

Among these methods, \cite{cozzolino2018camera} is the most similar one to PRNU-based methods as it estimates the ``noiseprint'' of a test image and the average one of some batched reference images using CNN, and computes the distance between two noiseprints to derive a heatmap for forgery localization. Besides, \cite{chen2017image} performs CRF (Camera Response Function) analysis through a CNN on edge patches and intensity gradient histogram (IGH) that carries information of CRF, while~\cite{mayer2018learned, mayer2019forensic, mayer2020exposing, ghosh2019spliceradar, bondi2017tampering} extract patch-level features and look for anomalies by training the network with patches labeled with the camera ID, given that tampered patches are usually captured by a different camera model and spliced with the original photo. Specifically, \cite{mayer2018learned} learns the similarity between pairs of patches to indicate whether these patches are from the same camera model, while~\cite{mayer2019forensic} further extends~\cite{mayer2018learned}'s work and scores the similarity of pairs of patches based on not only the camera source but also the editing operation and manipulation parameters. Furthermore, \cite{mayer2020exposing} proposes an improved method based on~\cite{mayer2018learned, mayer2019forensic} named forensic similarity graph to model the similarity between patches, in which patches are represented as vertices, whose edges are assigned based on their forensic similarity. Aside from the methods above, \cite{huh2018fighting} trains a CNN in a self-supervised manner using real images and photo EXIF metadata to learn the correlation between an image and its camera model. To be specific, the CNN is fed two image sources and performs a binary classification on each of the EXIF entries to determine whether the two sources share the same EXIF information. Given a test image, the network indicates a spliced region if it discovers an image patch leading to inconsistent EXIF predictions.

\section{Deepfake Detection}
\label{sec:deepfake_detection}
Face manipulation techniques~\cite{deepfakes,Thies_2016_CVPR,wu2018reenactgan,nirkin2019fsgan}, also known as Deepfakes, give rise to widespread concerns on the malicious tampering of human faces. This demands effective approaches that can detect photo-realistic forged faces generated by advanced Deepfake methods automatically. In this section, we first introduce the recent progress in Deepfake creation (Sec.~\ref{subsec:deepfake_generation}), and then summarize the existing public face forgery datasets (Sec.~\ref{subsec:deepfake_datasets}), finally, we will focus on the detection methods to fight against face manipulation (Sec.~\ref{subsec:deepfake_detection}).

\subsection{Deepfake Generation}
\label{subsec:deepfake_generation}
\begin{figure}[t]
  \centering
  \includegraphics[width=\linewidth]{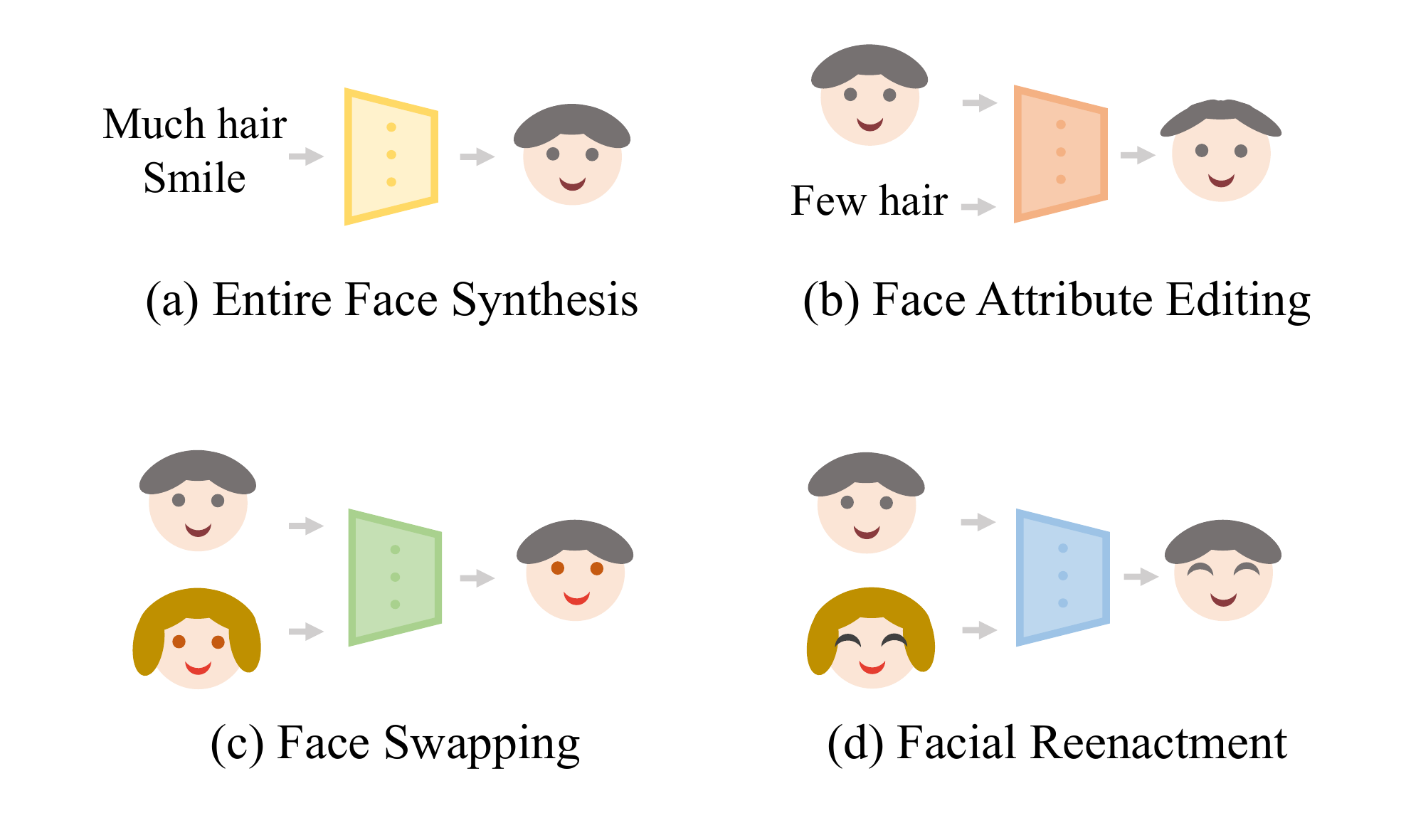}
  \caption{Illustration of four Deepfake synthesis techniques.}
  \label{fig:deepfake_generation}
\end{figure}

Human face analysis is a well-studied field in computer vision, which relates to various applications. As a result, existing analysis methods lay the foundation for the development of face manipulation techniques~\cite{natsume2018rsgan,nirkin2019fsgan,huang2020learning}. Specifically, based on the target and level of manipulation, Deepfake methods could be classified into four categories:  face synthesis (Sec.~\ref{subsubsec:entire_face_synthesis}), face attribute editing (Sec.~\ref{subsubsec:face_attribute_editing}), face swapping (Sec.~\ref{subsubsec:face_swapping}), and facial reenactment (Sec.~\ref{subsubsec:facial_reenactment}). We illustrate the process of them in Figure~\ref{fig:deepfake_generation} briefly.

\subsubsection{Face Synthesis}
\label{subsubsec:entire_face_synthesis}
Built upon the powerful generative models like GANs~\cite{goodfellow2014generative}, face synthesis methods~\cite{bao2018towards,shen2018faceid,yang2020one,deng2020disentangled} create non-existent faces with natural appearances and even realistic textures. ProGAN~\cite{karras2018progressive} proposes to train both the generator and discriminator progressively by gradually deepening the networks, which however still lacks the control over the style of the generated images. To address this issue, the StyleGAN family~\cite{karras2019style,karras2020analyzing,karras2020training,zhu2020improved,karras2021alias,wu2021stylespace} re-designs the generator architecture to enable the intuitive and scale-specific control of the synthesized results, inspired by the style transfer literature~\cite{huang2017arbitrary}. In addition, a group of work has also explored the use of multimodal information or structural priors to guide face generation, including landmarks~\cite{sun2022landmarkgan}, sketch~\cite{zhao2019generating,chao2019high,yu2020toward,chen2020deepfacedrawing,lin2020identity,li2020deepfacepencil,li2021sketch}, and voice~\cite{oh2019speech2face,wen2019face,duarte2019wav2pix}.

Compared with static image synthesis, video generation~\cite{vondrick2016generating} is much more challenging due to the fact that both appearance learning and dynamic modeling are required to create visually natural videos of human faces. To address this issue, video generation methods~\cite{vondrick2016generating,siarohin2019animating} typically learn the content and motion within videos in a decoupled manner, either with separate image and video generators~\cite{saito2017temporal,wang2020g3an,wang2020imaginator,tian2021a}, or by sampling from separate latent spaces~\cite{tulyakov2018mocogan,yu2022generating}. 

\subsubsection{Face Attribute Editing}
\label{subsubsec:face_attribute_editing}
Instead of synthesizing the complete faces, face attribute editing~\cite{Perarnau2016,natsume2018rsgan,zhang2018sparsely,chen2018facelet} aims to manipulate specific attributes, \eg, gender and age, of human faces, which therefore can be seen as fine-grained face manipulation. Denoting face attributes as different ``domains'', a series of methods~\cite{choi2018stargan,wu2019relgan,choi2020stargan} define face attribute editing as a multi-domain image-to-image translation task. These approaches~\cite{choi2018stargan,he2019attgan} take the domain labels, \eg, binary vectors, as inputs to translate the images into the target domain. However, \cite{liu2019stgan} argues that utilizing the target attribution vector is redundant and may even lead to poor results; instead, they directly adopt the difference attribute vectors to provide more valuable guidance. Unlike the above methods which reply on explicit domain labels, other work manipulates the latent representations in 2D~\cite{chen2019semantic,shoshan2021gan,abdal2021styleflow,shi2022semanticstylegan,he2021disentangled,xu2022transeditor} or 3D space~\cite{sun2022fenerf,medin2022most} to implicitly control the editing process, leveraging the entangled nature of the GAN latent space. 

In addition, the interactive face attribute manipulation has also attracted emerging attention due to its broad application prospects. With more user-friendly inputs, like texts~\cite{jiang2021talk,patashnik2021styleclip,xia2021tedigan,wei2022hairclip}, masks~\cite{dekel2018sparse,jo2019sc,lee2020maskgan,durall2021facialgan}, sketches~\cite{zeng2022sketchedit}, and color-histograms~\cite{afifi2021histogan}, the semantic manipulation of facial images could be achieved more flexibly.

\subsubsection{Face Swapping}
\label{subsubsec:face_swapping}
Face swapping~\cite{blanz2004exchanging,pierrard2008skin,korshunova2017fast,shen2017learning,nirkin2018face}, also known as face replacement or identity swapping, is another type of fine-grained face manipulation, which replaces the face of one person in an image with the face of another person. Typically, face swapping methods could be divided into two categories: classical computer graphics techniques~\cite{blanz2004exchanging,pierrard2008skin,bitouk2008face,dale2011video,garrido2014automatic,kemelmacher2016transfiguring,faceswap} and deep generative models~\cite{yan2018video,natsume2018fsnet,natsume2018rsgan,kim2022smooth}, the latter of which has gained considerate attention recently for its simpler pipeline and better results. Early methods~\cite{deepfakes,korshunova2017fast,nirkin2018face} require training the model separately for each pair of subjects, which will result in expensive training cost and thus limiting their potential applications. To tackle this problem, several works have developed subject-agnostic methods~\cite{nirkin2019fsgan,li2020advancing} that can perform facial identity swapping on arbitrary faces with one unified model by decomposing the identity information from the remaining attributes. In addition, developing efficient face swapping frameworks~\cite{chen2020simswap,xu2022mobilefaceswap} to enable the deployment on mobile devices, one-shot face swapping~\cite{zhu2021one}, and 3D-based facial replacement~\cite{moniz2018unsupervised,sun2018hybrid,wang2021hififace} have also drawn emerging concern in the community.

\begin{table*}[t]
\renewcommand\arraystretch{1.1}
\centering
  \caption{Statistics of  existing datasets for Deepfake detection.}
  \label{tab:deepfake_dataset}
  \setlength{\tabcolsep}{0.pt}
    \begin{tabular*}{\linewidth}{@{\extracolsep{\fill}}*{10}lc@{}}
    \toprule
    \multirow{2}{*}{\textbf{Generation}} &&
    \multirow{2}{*}{\textbf{Dataset}} && \multicolumn{2}{c}{\textbf{Real}} && \multicolumn{2}{c}{\textbf{Forged}} && \multirow{2}{*}{\textbf{\makecell[c]{Generation \\ Approaches}}} \\
    ~ && ~ && Video & Frame && Video & Frame & ~ \\
    \cmidrule{1-1} \cmidrule{3-3} \cmidrule{5-6} \cmidrule{8-9} \cmidrule{11-11} 
    \multirow{3}*{First} && UADFV~\cite{yang2019head} && 49 & 17.3k && 49 & 17.3k && 1 \\
    ~ && DF-TIMIT-LQ~\cite{korshunov2018deepfakes} && 320 & 34.0k && 320 & 34.0k && 2 \\
    ~ && DF-TIMIT-HQ~\cite{korshunov2018deepfakes} && 320 & 34.0k && 320 & 34.0k && 2\\
    \cmidrule{1-1} \cmidrule{3-11}
    \multirow{4}*{Second} && DFD~\cite{dfd} && 363 & 315.4k && 3,068 & 2,242.7k && 5\\
    ~ && Celeb-DF~\cite{li2020celeb}  && 590 & 225.4k && 5,639 & 2,116.8k && 1 \\
    ~ && FF++~\cite{rossler2019faceforensics++} && 1,000 & 509.9k && 4000 & 1,830.1k && 4 \\
    \cmidrule{1-1} \cmidrule{3-11}
    \multirow{4}*{Third} && DFFD~\cite{dang2020detection} && 1,000 &  58,703 && 3,000 & 240,336 && 7 \\
    ~ && DeeperForensics-1.0~\cite{jiang2020deeperforensics} && 1,000 & - && 5,000 & - && 5 \\
    ~ && DFDC~\cite{dolhansky2020deepfake} && 23,564 & - &&  104,500 & - && 8 \\
    ~ && ForgeryNet~\cite{he2021forgerynet} && 99,630 & 1,438.2k && 121,617& 1,457.9k && 15 \\
  \bottomrule
  \end{tabular*}
\end{table*}

\subsubsection{Facial Reenactment}
\label{subsubsec:facial_reenactment}
Facial reenactment~\cite{thies2015real,averbuch2017bringing} transfers the expression from the source person to the target person while preserving the identity information. Face2Face~\cite{thies2016face2face}, one of the most prominent graphics-based expression manipulation methods~\cite{thies2015real,averbuch2017bringing,kim2018deep,thies2018headon,thies2019deferred}, combines 3D model reconstruction and image-based rendering techniques for real-time face reenactment. Recently, the development of generative networks~\cite{goodfellow2014generative} has stimulated the progress of deep learning-based techniques~\cite{nirkin2019fsgan,qian2019make,lee2020maskgan,chen2020puppeteergan}. Early methods~\cite{zhu2017unpaired,choi2018stargan,wu2018reenactgan,huang2020learning,burkov2020neural,yao2021one} adopt simple encoder-decoder architectures for facial reenactment, which however suffers from poor transfer performance and prominent visual artifacts since facial expressions are subtle. It is widely-known that landmarks are neat, sufficient, and robust to reflect the structure of human faces~\cite{wang2022ft}, inspired by this, ~\cite{song2018geometry,hao2020far,otberdout2020dynamic,ha2020marionette,zakharov2020fast,liu2021li} apply driving landmarks to explicitly guide the reenactment process. Although decent synthesis results could be achieved with the above methods, training on a specific domain limits their capability to generate human faces with arbitrary expressions~\cite{pumarola2018ganimation}. In order to deal with this problem, a collection of work has explored the manipulation of expressions in a continuous space to achieve more natural reenactment results, with action units~\cite{pumarola2018ganimation,tripathy2020icface,tripathy2021facegan}, expression codes~\cite{ding2018exprgan}, or motion field~\cite{siarohin2019first}.

Apart from manipulating the facial expression based on driving faces, talking head generation~\cite{kumar2017obamanet,suwajanakorn2017synthesizing,wiles2018x2face,zhou2021pose} could be regarded as a special form of facial reenactment methods, which synthesizes a talking head video synchronized with a given audio clip. Resolving such a problem is highly challenging because it is non-trivial to faithfully relate the audio signals and face deformations, including expressions and lip motions~\cite{guo2021ad}. To mitigate this problem, most existing approaches achieve audio-head alignment by estimating the intermediate facial representations, \eg, 3D face shapes~\cite{karras2017audio,linsen2020ebt,thies2020neural,zhang2021flow}, dynamic kernels~\cite{ye2022audio}, action units~\cite{chen2021talking}, dense motion fields~\cite{wang2022one}, and landmarks~\cite{kumar2017obamanet,zakharov2019few,chen2019hierarchical,wang2020mead}. However, such multi-step transformation will inevitably lead to a loss of information and further affect the synthesis results. In contrast, ~\cite{guo2021ad} directly map the extracted audio features to dynamic neural radiance fields~\cite{mildenhall2020nerf,gafni2021dynamic} for motion estimation, and finally synthesize the high-fidelity talking head using volume rendering. 

\begin{figure}[t]
  \centering
  \includegraphics[width=\linewidth]{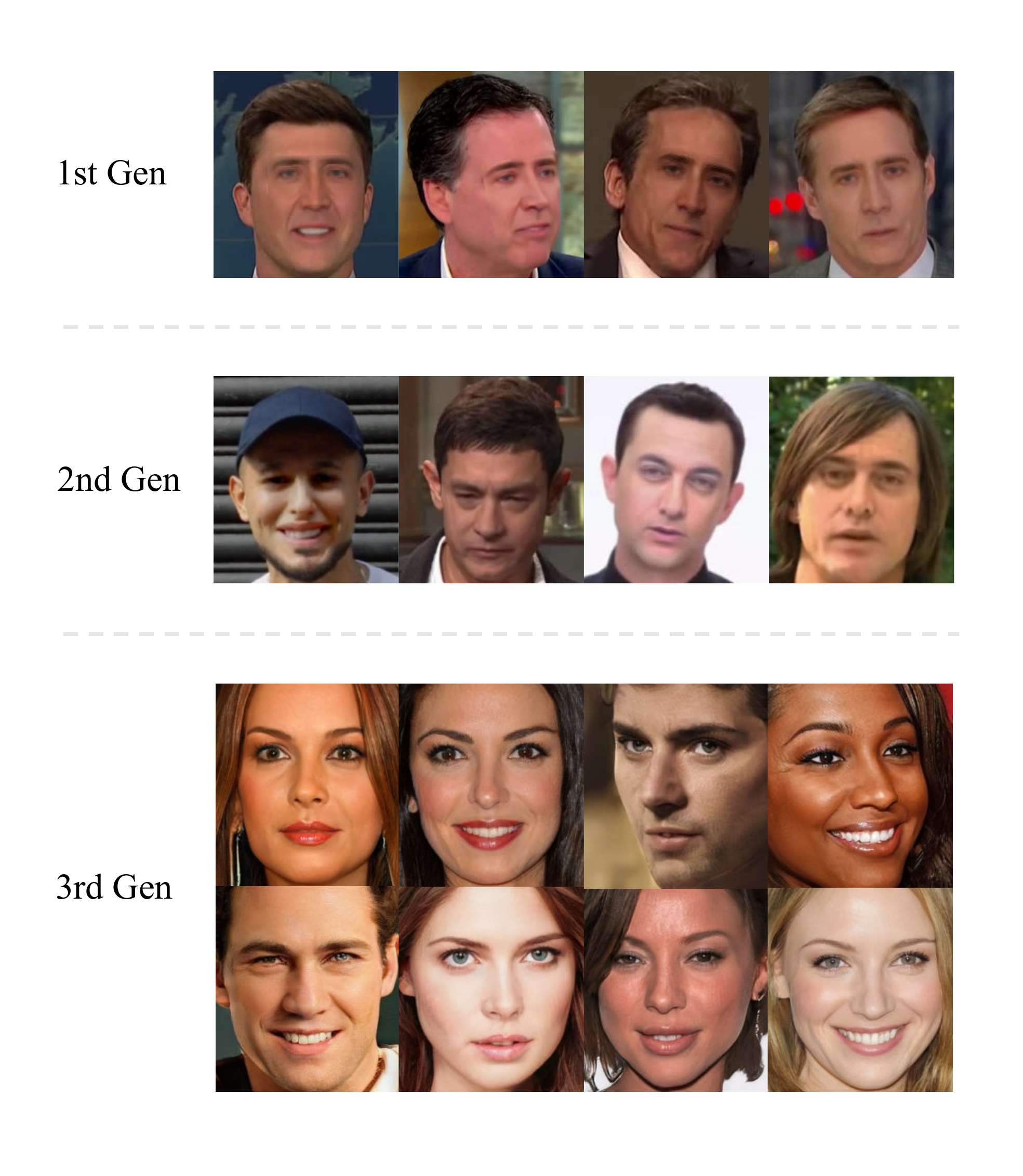}
  \caption{Comparisons between the Deepfake datasets of different generations. From top to down, we separately display the examples from UADFV, CelebDF, and DFFD.}
  \label{fig:deepfake_datasets}
\end{figure}

\subsection{Deepfake Datasets}
\label{subsec:deepfake_datasets}
The development and evaluation of face manipulation detection methods require large-scale datasets. In this section, we describe existing DeepFake datasets which are widely used to benchmark the effectiveness of different Deepfake detection methods. Based on the scale of forgery data and the generation techniques, these datasets could be divided into three generations.

\subsubsection{First-generation Datasets}
The first-generation datasets are relatively small-scale datasets generated with open-source Deepfake creation tools, including UADFV~\cite{yang2019head} and DeepFake-TIMIT~\cite{korshunov2018deepfakes}. \textbf{UADFV} dataset~\cite{yang2019head} contains 49 real videos collected from YouTube and 49 Deepfake videos that are generated using FakeAPP~\cite{fakeapp}. \textbf{DeepFake-TIMIT}~\cite{korshunov2018deepfakes} includes 320 real videos and 640 Deepfake videos (320 high-quality and 320 low-quality) generated with faceswap-GAN~\cite{faceswapgan}. 
    
\subsubsection{Second-generation Datasets}
The second-generation Deepfake datasets consist of DFD~\cite{dfd}, Celeb-DF~\cite{li2020celeb}, and FF++~\cite{rossler2019faceforensics++}, which are the most widely adopted datasets by far. The Google/Jigsaw \textbf{DeepFake Detection dataset}~\cite{dfd} (DFD) includes 3,068 Deepfake videos that are generated based on 363 original videos. The \textbf{Celeb-DF} dataset~\cite{li2020celeb} contains 590 real videos and 5,639 Deepfake videos created using the same synthesis algorithm. The \textbf{FaceForensics++} (FF++) dataset~\cite{rossler2019faceforensics++} has 1,000 real videos from YouTube and 4,000 corresponding Deefake videos that are generated with 4 face manipulation methods: Deepfakes~\cite{deepfakes}, FaceSwap~\cite{faceswap}, Face2Face~\cite{Thies_2016_CVPR}, and NeuralTextures~\cite{thies2019deferred}. 

\subsubsection{Third-generation Datasets}
The most recent datasets, including DFFD~\cite{dang2020detection}, DeeperForensics-1.0~\cite{jiang2020deeperforensics}, DFDC~\cite{dolhansky2020deepfake}, and ForgeryNet~\cite{he2021forgerynet}, are typically regarded as the third-generation datasets. \textbf{Diverse Fake Face Dataset}~\cite{dang2020detection}, also termed as DFFD, adopts the images from FFHQ~\cite{karras2019style} and
CelebA~\cite{liu2015deep} datasets source subset, and synthesizes forged images with various Deepfake generation methods. DeeperForensics-1.0~\cite{jiang2020deeperforensics} consists of 60,000 carefully-collected videos with a total of 17.6 million frames. It is worth mentioning that extensive real-world perturbations are applied on the generated images to further expand the scale and diversity. The Facebook \textbf{DeepFake Detection Challenge dataset}~\cite{dolhansky2020deepfake} (DFDC) is part of the DeepFake detection challenge, which has 1,131 original videos and 4,113 Deepfake videos. The \textbf{ForgeryNet}~\cite{he2021forgerynet} is currently the largest publicly available Deepfake dataset, which contains 2.9 million images and 221,247 videos. The forged images in ForgeryNet~\cite{he2021forgerynet} are generated with 7 image-level approaches and 8 video-level approaches, from which four forgery identification tasks are derived: 1) image forgery classification; 2) spatial forgery localization; 3) video forgery classification; 4) temporal forgery localization. 

We summarize the statistics of these existing Deepfake datasets in Table~\ref{tab:deepfake_dataset}. Examples from the Deepfake datasets of different generations are displayed in Figure~\ref{fig:deepfake_datasets}.

\subsection{Deepfake Detection}
\label{subsec:deepfake_detection}
With the prominent advances in Deepfake generation techniques, the synthesized images are becoming more photo-realistic, making it extremely difficult to distinguish fake faces even for the human eyes. At the same time, these forged images might be distributed on the Internet for malicious purposes,
which could bring societal implications. 

As a response to the increasing concern of the above challenges, various forensic methods~\cite{zhou2017two,afchar2018mesonet,nguyen2019multi} are proposed, which typically take as inputs a face region cropped out of an entire image and produce a binary real/fake prediction. In the section below, we present a comprehensive discussion of existing Deepfake detection methods, where we group them into two major categories: image detection models (Sec.~\ref{subsubsec:image_detection}) and video detection models (Sec.~\ref{subsubsec:video_detection}). 

\subsubsection{Deepfake Image Detection}
\label{subsubsec:image_detection}
Manipulating face images will inevitably leave some tampering clues, \eg, handcrafted features, GAN fingerprints, and deep visual features, which, can be adopted by defense models for Deepfake image detection~\cite{nguyen2019capsule}.

\fakeparagraph{Handcrafted features-based methods} typically take a closer look at the inconsistencies produced by the synthesis process of Deepfakes for face forensics. Color space is widely adopted by forgery discrimination methods due to its robustness to various post-precessing, \eg, resizing and compression. Specifically, based on the observation that Deepfake images are significantly different from real ones in the chrominance components of HSV and YCbCr color spaces, \cite{li2020identification} introduces a color statistics-based feature set to identify the forged faces. \cite{he2019detection} further extends this to Lab color space and concatenate the deep representations from different color spaces to obtain final detection results. Moreover, taking into consideration the consistent relationships among different color channels, \cite{mccloskey2019detecting} identifies the Deepfake images through the frequency of over-exposed pixels, while ~\cite{barni2020cnn,goebel2021detection} distinguishes GAN-generated images from camera-generated images by calculating cross-band co-occurrence matrices and spatial co-occurrence matrices. 

In addition, considering it is difficult for deep networks to replicate the high frequency details in real images, \cite{bai2020fake} combines the statistical, oriented gradient and blob features in the frequency domain for fake painting detection. Besides, the lack of global constraints may also lead to forged faces with unreasonable head poses, which inspires a group of work to distinguish GAN synthesized fake faces with detected landmarks~\cite{yang2019exposing} or estimated 3D head models~\cite{yang2019head}. Other than these, \cite{li2019exposing} detects the warping artifacts resulted from transforming the synthetic face regions to the pristine images, which achieves a competitive performance with a limited amount of fake data. \cite{koopman2018detection} explores the use of photo response non uniformity (PRNU) analysis to identify Deepfakes. \cite{chen2021defakehop} derives the rich face representation using PixelHop++ units~\cite{chen2020pixelhop++}, and finally adopts the ensemble of different regions and frames for classification. ~\cite{fernandes2020detecting} detects the Deepfake videos using the state-of-the-art attribution based confidence (ABC) metric, which does not require the access to the training data.

\fakeparagraph{GAN fingerprints-based methods} explore the marks that are commonly shared by GAN-generated images~\cite{yu2019attributing} for forgery detection. Early methods adopt unsupervised machine learning techniques to discover the GAN-specific features. \cite{guarnera2020deepfake} first extracts a set of local features with an Expectation Maximization (EM) algorithm to model the convolutional traces, which are input to naive classifiers to discriminate between authentic and synthesized images. In order to improve the robustness towards various attacks on images such as JPEG Compression, mirroring, rotation, \cite{giudice2021fighting} transforms the suspicious images to frequency domain through Discrete Cosine Transform (DCT) and examines the statistics of the DCT coefficients for forgery detection. The above methods capture the GAN fingerprints at a single scale, which, therefore, cannot achieve competitive results on high-quality fake images. To mitigate this issue, ~\cite{ding2021does} employs a hierarchical Bayesian approach for multi-scale latent GAN fingerprint modeling. 

Recent work has attempted to employ deep neural networks (DNN)~\cite{yang2022deepfake} to identify the subtle artifacts left by GAN. \cite{zhang2019detecting,frank2020leveraging,huang2022fakelocator} study the unique artifacts induced by the up-sampling of GAN pipelines to develop robust GAN image classifiers. \cite{yu2019attributing} discovers that different training strategies will leave distinctive fingerprints over all generated images, which can be utilized to enable fine-grained image attribution and even model attribution. \cite{yang2022deepfake} explores the existence and properties of globally consistent GAN fingerprints through empirical study. Based on this, they develop a simple yet effective approach to capture GAN artifacts by pre-training on image transformation classification and patchwise contrastive learning.

\begin{figure}[t]
\centering
\subfloat[Deepfake]{\includegraphics[width=.68in,height=2.5in]{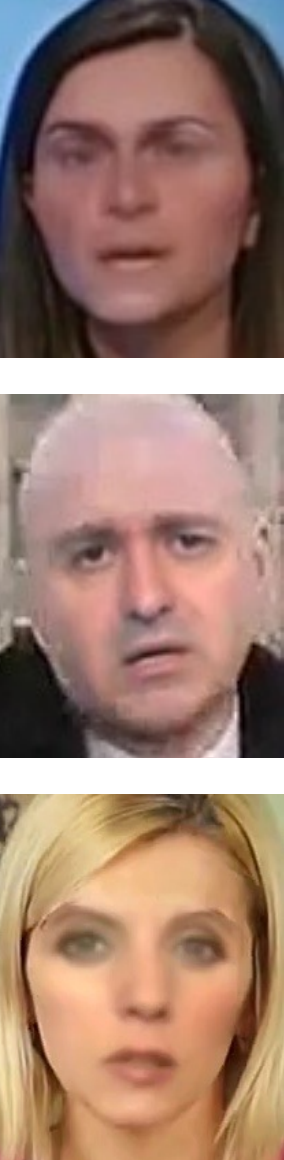}}
\hfill
\subfloat[Mask]{\includegraphics[width=.68in,height=2.5in]{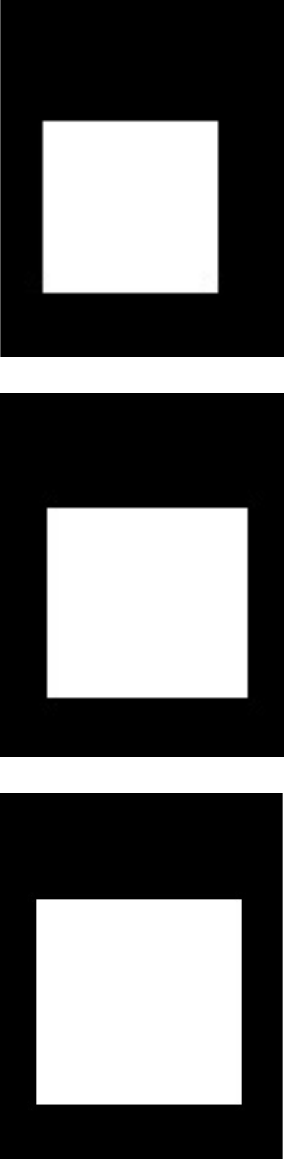}}
\hfill
\subfloat[Xcep~\cite{chollet2017xception}]{\includegraphics[width=.68in,height=2.5in]{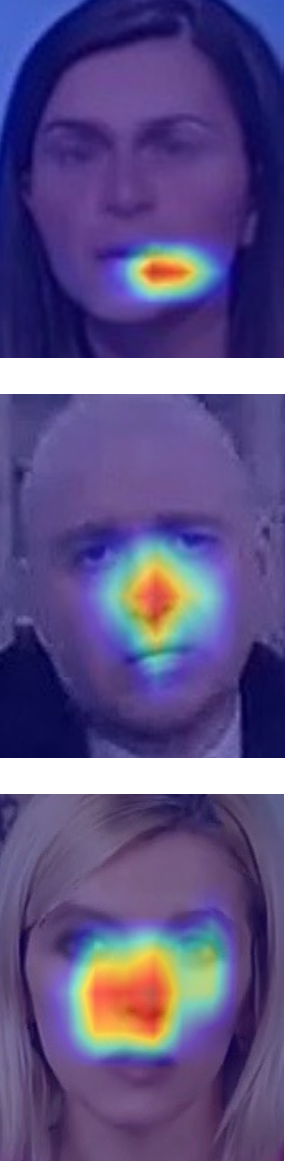}}
\hfill
\subfloat[EN~\cite{tan2019efficientnet}]{\includegraphics[width=.68in,height=2.5in]{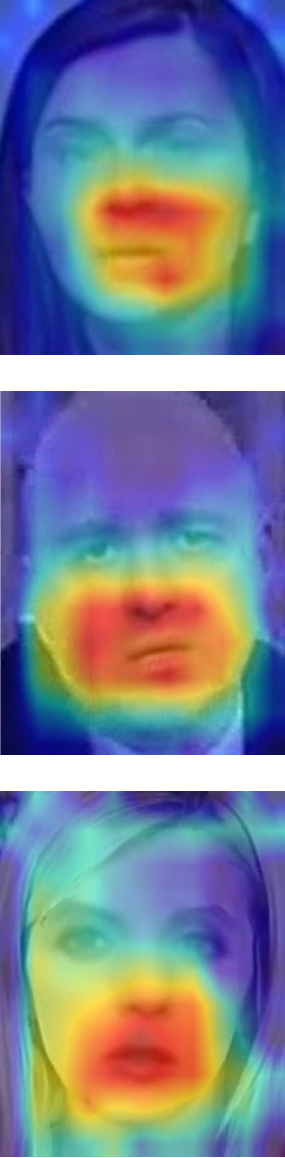}}
\hfill
\subfloat[Gram~\cite{liu2020global}]{\includegraphics[width=.68in,height=2.5in]{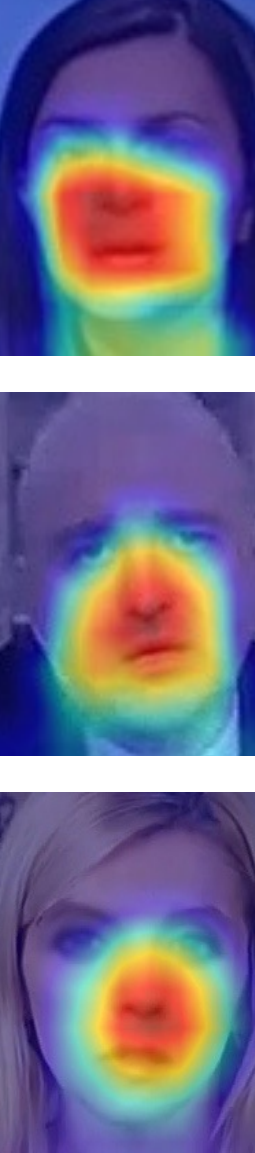}}
\hfill
\caption{Visualization of several deep features-based Deepfake detection models using Grad-CAM. Xcep denotes Xception~\cite{chollet2017xception}, EN denotes EfficientNet~\cite{tan2019efficientnet}, Gram denotes GramNet~\cite{liu2020global}.}
\label{fig:gradcam}
\end{figure}

\fakeparagraph{Deep features-based methods} rely on deep models to mitigate the security threat brought by Deepfakes. Pioneering work~\cite{nguyen2019capsule,guo2021fake} typically captures artifacts from the face regions with stacked convolutional operations. More specifically, \cite{zhou2017two} detects tampered face images with a two-stream network, where a face classification stream captures the forgery artifacts and a patch triplet stream recognizes noise residual evidence. \cite{gandhi2020adversarial,khan2021adversarially} improve the robustness of Deepfake detectors with Lipschitz regularization and model fusion, separately. In addition, considering binary classification tends to result in easily overfitted models, \cite{nguyen2019multi,li2019zooming,kong2022detect} propose to locate the manipulated regions to mitigate overfitting. Beyond these, the growing concerns towards Deepfakes have promoted the emergence of more advanced forgery detection approaches, as will be discussed in the following.  We visualize the feature map of several Deepfake detection models with Grad-CAM~\cite{selvaraju2017grad} in Figure~\ref{fig:gradcam}.

\vspace{1em}
\textit{a) Contrastive-based approaches.} Deepfake forensics is essentially a discriminative problem, \ie, fitting the decision boundary on large-scale datasets. Therefore, focusing solely on the differences between binary categories will lead to limited generalization in unseen domains~\cite{sun2022dual}. With this in mind, a variety of methods~\cite{hsu2018learning,feng2020deep,kumar2020detecting,fung2021deepfakeucl} seek to expand the discrepancies between authentic and fake images in feature space by means of contrastive learning. \cite{hsu2018learning} learns joint discriminative features from heterogeneous training samples, \ie, fake images generated with different GAN models, with a vanilla contrastive loss. In order to improve the robustness towards image compression algorithms, \cite{cao2021metric} encodes paired original and compressed forgeries to a compression-insensitive embedding feature space, in which the distance between genuine and forgery data are maximized while the distance between paired images are reduced through the supervision of a metric loss. The instance-level contrastive learning models the association between different images or the same image under different compression levels, which is limited to learn coarse-grained representations. Comparatively, \cite{sun2022dual} further introduces a intra-instance contrastive learning mechanism to mine deeper on the local inconsistencies within the forged faces.

\vspace{1em}
\textit{b) Attention-based approaches.}  Considering most Deepfake generation methods, \eg, expression reenactment, only manipulate parts of the face images, identifying the suspect regions and ``paying more attention to'' their visual contents could effectively improve the detection results. Inspired by this, a series of attention-based methods have been proposed. \cite{dang2020detection} applies Principal Component Analysis (PCA) to several ground-truth manipulation masks to obtain the base attention maps, which are dynamically weighted and summed to the mean face with a manipulation appearance modeling module. \cite{chen2021attentive} first segments the key fragments, \eg, eyes and mouths, from a face image, and then looks into these regions with local attention modeling. The methods mentioned above rely on manually-designed areas of interest, which limits the representative capability of Deepfake detectors. Comparatively, \cite{zhao2021multi} combines multiple spatial attention heads to automatically attend to different local parts with trainable convolution layers. Based on the observation that the reconstruction difference of real and fake faces follows significantly different distributions, \cite{cao2022end} uses a reconstruction-guided attention module to enhance the features derived from the multi-scale graph reasoning. In addition, \cite{wang2021representative} presents an attention-based data augmentation framework to encourage the detector to attend to subtle forgery traces. Specifically, they first generate a Forgery Attention Map (FAM) by calculating the gradients with respect to the classification loss, and then occlude the Top-N
sensitive facial regions intentionally to facilitate the detector to mine deeper into the regions ignored before.

\vspace{1em}
\textit{c) Local relation-based approaches.} Most existing Deepfake detection methods formulate face forgery detection as a classification problem with global supervisions, \eg,  binary labels or manipulated masks, which are insufficient to learn discriminative features and prone to overfitting~\cite{chen2021local} without explicit local relation modeling. To tackle this issue, \cite{liu2020global} enhances texture features with stacked Gram blocks. \cite{zhao2021learning} assumes a forged image would contain different source features at different positions, therefore, the forged images could be identified by measuring their self-consistency. Similarly, \cite{chen2021local} models the relations of local patches by calculating the similarity between features of different regions.

\vspace{1em}
\textit{d) Frequency-based approaches.}  The prominent advances in Deepfake techniques have made the visual artifacts in fake images more and more subtle and imperceptible. In addition, the common image processing algorithms, \eg, compression, will further contaminate the forgery clues in RGB domain. Fortunately, several prior studies suggest that these artifacts can be captured in frequency domain~\cite{durall2019unmasking,li2004live,wang2020cnn,yu2019attributing}, in the form of unusual frequency distributions when compared with real faces~\cite{qian2020thinking}. This motivates a sequence of work~\cite{liu2021spatial,li2021frequency,gu2022exploiting,woo2022add,jeong2022bihpf} to use frequency information as a complementary modality to develop stronger and more robust Deepfake detection methods. \cite{qian2020thinking} uses a two-stream collaborative framework by combing a Frequency-aware Decomposition stream to discover salient frequency bands and a Local Frequency Statistics stream to extract local frequency statistics. \cite{jeong2022bihpf} notices that the artifacts have large magnitudes in the high-frequency components and are oftentimes located in the surrounding background of the image rather than the central region, therefore, they combine the frequency-level High-Pass Filters (HPF) for amplifying the magnitudes of the artifacts in the high-frequency components, and the pixel-level HPF for emphasizing the pixel values in the surrounding background in the pixel domain. \cite{woo2022add} distills a teacher model pretrained on high-quality images with a novel frequency attention distillation to detect low-quality deepfakes. The aforementioned methods generally make use of the frequency information from a global perspective, \cite{gu2022exploiting} instead employs the patch-wise Discrete Fourier Transform~\cite{xu2020learning} with the sliding window to decompose the RGB input into fine-grained frequency components, considering the forgery clues are oftentimes hidden in the local patches.

The aforementioned approaches detect forgery artifacts with convolutional layers, which fail to model the consistency of pixels globally due to the limited size of receptive fields. To tackle this issue, \cite{wang2021m2tr} adopts the Transformer architecture~\cite{vaswani2017attention} to capture long-term dependencies between different face regions, and extract multi-scale tampering features by splitting the feature maps into patches of different sizes. Following~\cite{wang2021m2tr}, \cite{khormali2022dfdt} further selects multi-scale salient patches based on attention weights to  capture sensitive information. ~\cite{heo2021deepfake} distills the knowledge from EfficientNet~\cite{tan2019efficientnet}, the state-of-the-art model on the DFDC dataset, to fully boost the performance of vision transformer for Deepfake detection.

\begin{figure}[t]
  \centering
  \includegraphics[width=\linewidth]{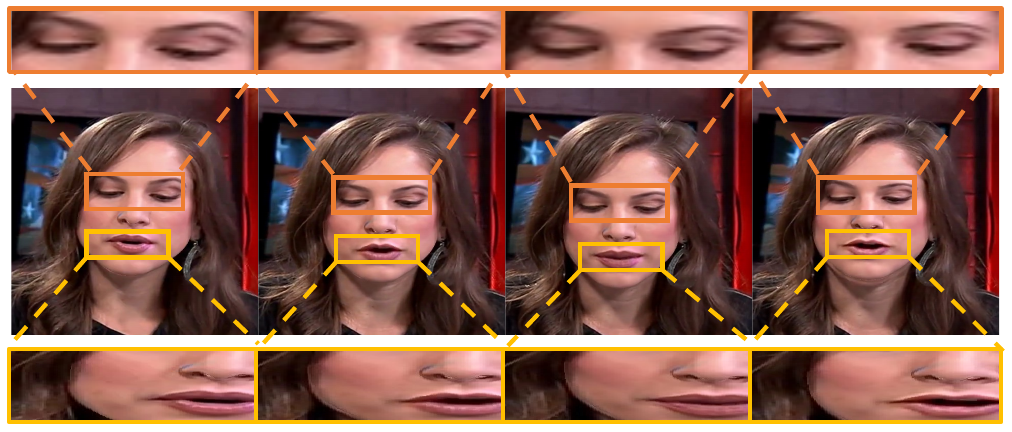}
  \caption{Illustration of Spatial-Temporal forgery clues in Deepfake videos. Both visual artifacts (spatially) and biological behaviors (temporally, \eg, abnormal blinking rate) can be exploited to detect Deepfake videos.}
  \label{fig:st_artifacts}
\end{figure}

\subsubsection{Deepfake Video Detection}
\label{subsubsec:video_detection}
Nowadays, forged faces that are circulating on the Internet take the form of videos, since videos can be more persuasive. Although image-level tampering detection methods can be directly applied to videos, they lack the capability to model temporal dynamics in videos. We visualize the spatial-temporal artifacts in Deepfake videos in Figure~\ref{fig:st_artifacts}. This has motivated a line of work on video-level tampering detection~\cite{bondi2020training,chen2020fsspotter}. According to the type of features of interest, we further classify the Deepfake video detection methods into biological behaviors-based approaches and spatial-temporal visual clues based approaches.

\vspace{1em}
\paragraph{Biological behaviors.} Although deep generative models can synthesize realistic faces, they oftentimes fail to reproduce some biological behaviors intrinsic to human beings. This inspires lots of work~\cite{ciftci2020fakecatcher} to detect forged videos through the analysis of biological behaviors. Eye blinking is a widely used clue by forensics methods~\cite{li2018ictu,jung2020deepvision} since it is a voluntary and spontaneous action that does not require conscious effort for humans~\cite{jung2020deepvision}. Specifically, \cite{li2018ictu} discovers the absence of faces with eye closed in the training datasets results in the lack of eye blinking in forged videos. To this end, they combine a convolutional neural network with a recursive neural network (RNN) to capture the phenomenological and temporal regularities of eye blinking for forgery detection. \cite{jung2020deepvision} analyzes the pattern of eye blinking in more detail, \ie, the period, repeated number, and elapsed eye blink time when eye blinks were continuously repeated within a very short period of time. 

Some other appearance cues like lip movements~\cite{agarwal2020detecting,yang2020preventing,lin2021lip,haliassos2021lips} and head pose~\cite{yang2019head,agarwal2019protecting} have also gained widespread concern for Deepfake video detection. \cite{agarwal2020detecting} focuses on the visemes associated with words like M (mama), B (baba), or P (papa) in which the mouth must completely close to pronounce these phonemes but usually not the case in fake videos. \cite{haliassos2021lips} first pretrains a Visual Speech Recognition (VSR) network to learn the internal consistency between the shape of lips and spoken words, and then fine-tunes a temporal network on fixed mouth embeddings for video forgery detection. \cite{agarwal2019protecting} proposes a customized forensic method for specific individuals by modeling their distinct patterns of facial and head movements. \cite{yang2019head} observes that most Deepfake synthesis methods are incapable of matching the location of facial landmarks reasonably, which are invisible to human eyes but can be revealed from 3D head poses estimation. Inspired by this, they train a simple SVM classifier on the estimated head poses to differentiate original and fake videos.

Another interesting point is that several recent studies~\cite{ciftci2020fakecatcher,fernandes2019predicting,qi2020deeprhythm} present the heart rate estimated from visual contents can be utilized for face video forensics. \cite{fernandes2019predicting}, for instance, predicts the heart rate with a state-of-the-art Neural Ordinary Differential Equations (Neural-ODE) model, which could further facilitate the Deepfake detection by identifying abnormal heartbeat rhythms. Additionally, it is recognized that heartbeat will change the skin reflectance over time due to the hemoglobin content in the blood, which however are oftentimes disrupted or even entirely broken in forged videos. This inspires a line work~\cite{ciftci2020hearts,qi2020deeprhythm,hernandez2020deepfakeson} to make use of photoplethysmography (PPG) to identify Deepfake videos. Specifically, \cite{hernandez2020deepfakeson} separately input video frames and differences between adjacent frames to an appearance model and a motion model for Deepfake detection, which initializes the weights from an existing heart rate estimation framework to transfer its knowledge. \cite{qi2020deeprhythm}, from another point of view, incorporates both a heart rhythm motion amplification module and a spatial-temporal
attention module to expose subtle manipulation artifacts.

\vspace{1em}
\paragraph{Deep features.} The prominent success of deep learning makes it attractive to detect forged videos using end-to-end trainable deep networks~\cite{afchar2018mesonet,nguyen2019capsule,nguyen2019multi,li2020sharp}. Although current Deepfake techniques achieve impressive performance regarding quality and controllability, they typically manipulate images at frame-level, thus struggling to preserve temporal coherence~\cite{zheng2021exploring}. To this end, a series of work~\cite{zheng2021exploring,trinh2021interpretable} focus on uncovering the temporal discordance to identify forged videos. ~\cite{trinh2021interpretable} learns a set of prototypes to represent specific activation patterns in a patch of the convolutional feature maps, which further interact with a test video to makes predictions based on their similarities. \cite{wang2020exposing} presents a motion feature-based video forensic method, which calculates the correlation matrices from local motion features to represent the temporal smoothness of the whole video. \cite{zheng2021exploring} notices that some discontinuity in fake videos may happen in frames that are not in the neighborhood, therefore, they adopt a light-weight temporal Transformer to explore the long-term temporal coherence. \cite{hu2022finfer} solves the Deepfake detection problem from a different perspective, which predicts the facial representations of future frames using an autoregressive model and distinguishes the fake videos according to the estimation error. The sparse sampling strategy is shared by most existing methods, which however disables them from modeling the local motions among adjacent frames. To counter this issue, ~\cite{gu2022delving} proposes a novel sampling unit named snippet which contains a few successive videos frames for local temporal inconsistency learning. 

Despite the favorable results achieved, ignoring spatial cues still limits the performance and scalability of the above temporal-focused approaches. Comparatively, a variety of work utilizes both spatial and temporal information via 2D/3D CNN~\cite{zhang2021detecting} or Conv-RNN architectures~\cite{guera2018deepfake,masi2020two}. Among them, a considerable amount of work~\cite{matern2019exploiting,tursman2020towards,gu2021spatiotemporal} focuses on capturing the spatial-temporal inconsistency of suspicious videos. Specifically, ~\cite{gu2021spatiotemporal,hu2021dynamic} adopt a two-branch network to separately attend to the intra-frame and inter-frame inconsistencies. ~\cite{dong2020identity} presents an identity-driven face forgery detection approaches, which requires the access to both a suspect image and a reference image indicating the target identity information, and outputs a decision on whether the two faces share the same identities. 

\vspace{1em}
\paragraph{Multimodal information.} Videos are embedded with multimodal information, \ie, motion (optical flow), and audio by nature, which inspires several works to explore multimodal inputs~\cite{amerini2019deepfake,chintha2020leveraging,chintha2020leveraging} or cross-modal alignment~\cite{chugh2020not,zhou2021joint,fernando2019exploiting} for Deepfake forensics. \cite{amerini2019deepfake} simply calculates the optical flow fields in an off-line manner, which are further input a CNN classifier to discover the inter-frame dissimilarities. In addition, the audio-visual synchronization is an intrinsic property of pristine videos, therefore, manipulation of either modality will result in non-negligible dissonance, \ie, loss of lip-sync, unnatural facial and lip movements, \etal~\cite{zhou2021joint}. Based on this, ~\cite{chugh2020not} learns discriminative representations for the audio and visual inputs in a chunk-wise manner, employing the cross-entropy loss for individual modalities, and a contrastive loss that models inter-modality similarity. \cite{mittal2020emotions} adopts a Siamese network-based architecture to separately extract the features of video and audio modalities, and constrain the consistency learning by a triplet loss function. ~\cite{haliassos2022leveraging} first learns the ``temporally dense'' video features, \eg, facial movements, expression, and identity, in a self-supervised cross-modal manner, and then fine-tunes a binary forgery classifier on the learned representations.

\subsubsection{Generalizable Deepfake Detection}
\label{subsubsec:generalizable_deepfake_detection}
The Deepfake generation techniques are evolving rapidly and continuously, therefore, the development of face forensic methods with distinguished generalization capabilities has gained sustained attention~\cite{hulzebosch2020detecting,chai2020makes,sun2021domain,luo2021generalizing,chen2022self,yu2022improving}. However, the data-driven property makes most face forgery detection methods struggle to achieve competitive results in cross-database evaluations~\cite{luo2021generalizing}. In order to counter this challenge, ~\cite{li2020face} captures the visual discrepancies across the blending boundaries for face tampering detection, based on the observation that most Deepfake methods blend the manipulated face regions with background images. ~\cite{marra2019incremental} proposes an incremental learning method, aiming to classify Deepfakes generated by novel models without worsening the performance on the previous ones. ~\cite{khalid2020oc}, further formulates the Deepfake detection as a one-class anomaly detection problem, which trains a one-class Variational Autoencoder (VAE) only on real face images and detects fake images by treating them as anomalies. 

Defining the forged images generated by diversified Deepfake techniques as different domains, a series of face forgery forensic methods~\cite{cozzolino2018forensictransfer,pmlr-v119-jeon20a,sun2021domain,pmlr-v119-jeon20a,kim2021fretal} build upon domain adaptation for Deepfake detection by transferring the knowledge from the source (teacher) to target (student) domains. ~\cite{sun2021domain} divides the source domain into training domain and meta-domain to simulate the domain shift, and updates the model weights using a meta-optimization strategy. ~\cite{cozzolino2018forensictransfer} proposes a weakly-supervised setting, where only a handful
of training samples from target domain are available. They disentangle the decision space combining an activation loss and a reconstruction loss, which facilitate the detector to learn discriminative and complete representations respectively.

In order to encourage the Deep forensic models to learn the intrinsic features to classify the generated and real face images, several studies propose to apply preprocessing on the training data, \eg, Gaussian blur and Gaussian noise~\cite{wang2020cnn} and JPEG compression~\cite{wang2020cnn}, to eliminate noisy low level cues. Along the same lines, ~\cite{chen2022self} further makes use of the adversarial training strategy to dynamically synthesize both challenging and diversified forgeries.

\begin{figure}[t]
  \centering
  \includegraphics[width=\linewidth]{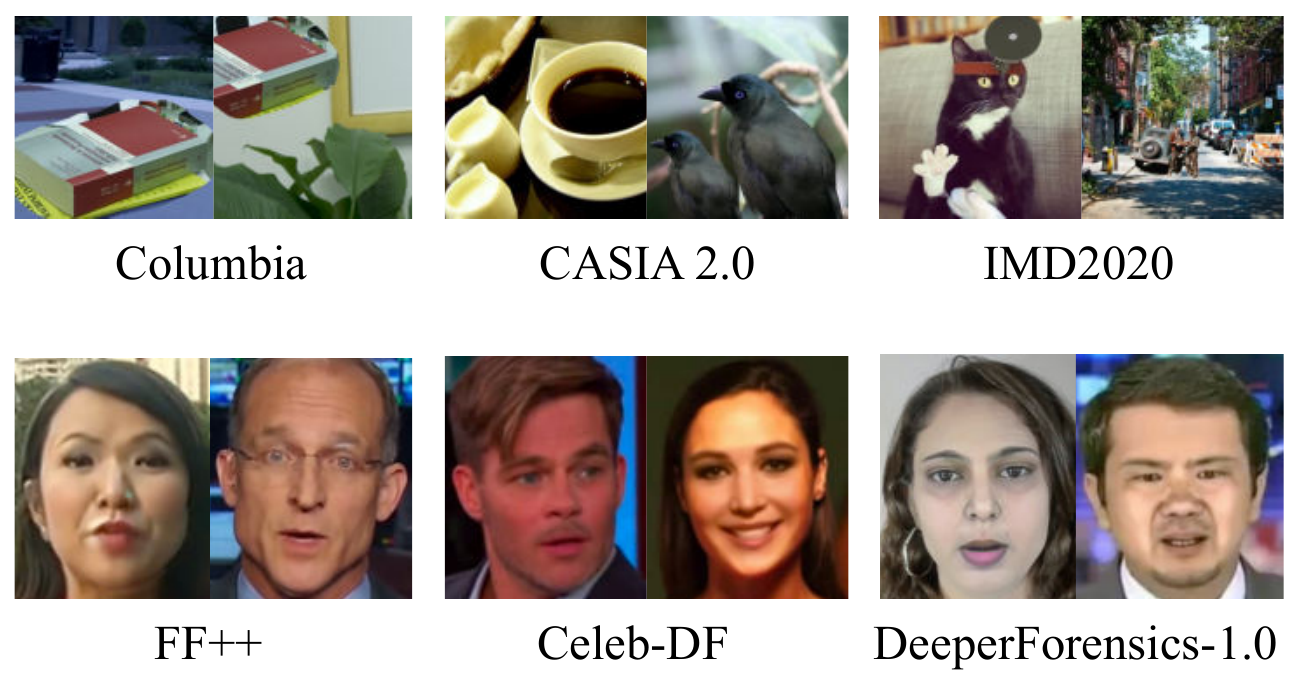}
  \caption{Illustration of the diversified forgery patterns in different tampering (top) and Deepfake (bottom) datasets.}
  \label{fig:generalization}
\end{figure}

\section{Challenges and Future Directions}
\label{sec:challenges}
Although recent years have witnessed significant progress in media forensics, there still exists a number of open problems that are worth exploring. It is worth mentioning that the common tug-of-war nature between generation methods and forensic methods makes both tampering detection and Deepfake detection face many shared problems. In this section, we summarize the current challenges together with possible future directions in this field.

\fakeparagraph{Generalization}: the blooming media manipulation techniques could generate more and more realistic and diversified forged data (see Figure~\ref{fig:generalization}), which requires the media forensic algorithms to develop favorable generalization ability, \ie, a model trained on a specific forgery methods should work well against another unknown one because potential manipulation types are typically unknown in the real-world scenarios. 

To mitigate this issue, ~\cite{zampoglou2015detecting,heller2018ps} collect a large quantity of forged images in the wild to construct a large-scale tampering datasets, which generally go through numerous manual manipulations to foster the generalization of data-driven detection methods. On the other side, several Deepfake detection methods~\cite{cozzolino2018forensictransfer,kim2021fretal} have also explored to alleviate this issue from the perspective of model design and training strategy (Sec.~\ref{subsubsec:generalizable_deepfake_detection}), but more work needs to be done in both tampering detection and Deepfake detection.

\fakeparagraph{Robustness}: multimedia data, \eg, images and videos circulating on the Internet are oftentimes processed, \eg, compression, resizing, and Gaussian blurring, during upload, transmission and download, which however would inevitably destroy the details in pristine data and thus bringing more severe challenges to forgery detection. In addition, malicious users even deliberately impose invisible perturbations on the forged data to fool the detection tools.

\begin{figure}[t]
  \centering
  \includegraphics[width=\linewidth]{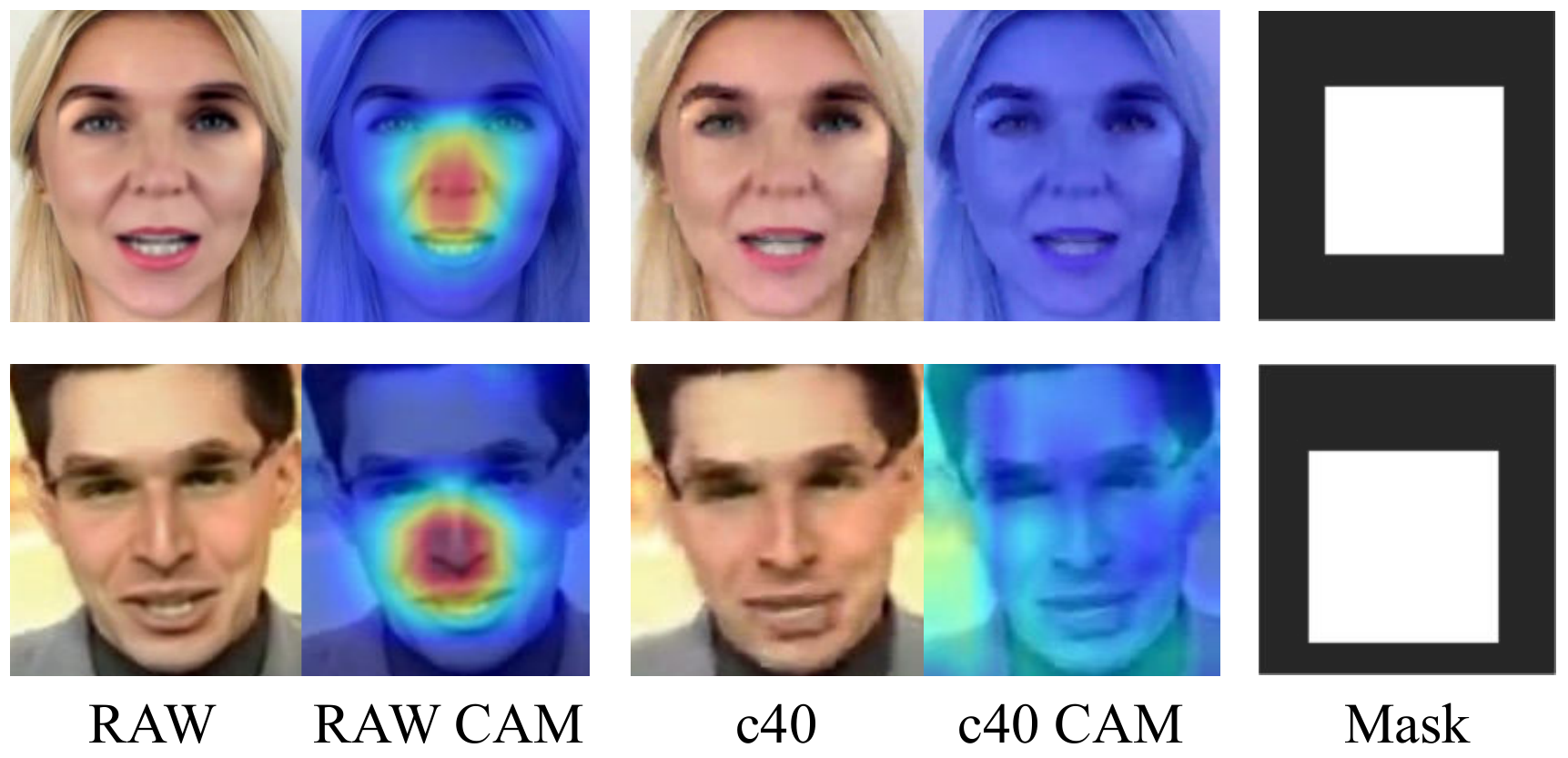}
  \caption{Visualization of the Grad-CAM on different subsets of FF++. From left to right, we display the results on raw and c40 (compressed with a rate factor of 40) images using the model pretrained on the raw subset of FF++.}
  \label{fig:robustness}
\end{figure}

However, forensic models trained on high-quality images oftentimes suffer from severe performance degradation on post-processed images. For instance, we visualize the Grad-CAM on the raw fake images and the corresponding compressed images using the model pretrained on raw images in Figure~\ref{fig:robustness}. The results demonstrate that although compression algorithms will not lead to a significant visual difference, the detector still fails to work. From a defensive standpoint, nevertheless, a ``good'' manipulation detector should be robust towards the common distortion algorithms. A good case in point is~\cite{wu2022robust}, as it thoroughly analyzes the impact of Online Social Networks on tampered images and offers a solution. Since there is generally no standard of handling images in web applications for detectors, detectors themselves must be robust enough against these various scenarios.

\fakeparagraph{Model Attribution}: most existing forensics methods typically focus on binary classification \ie, real/fake. Nevertheless, attributing the manipulated images/videos to the model generating them is also of prominent significance for legal accountability and intellectual
property protection purposes~\cite{girish2021towards}. Although several approaches~\cite{yu2021artificial,yang2022deepfake} have explored to perform attribution on Deepfake images for multiple GAN architectures, the problem of model attribution for both tampering and Deepfake detection is far from studied and solved sufficiently.

\fakeparagraph{Multimodal information}: with the advancement of tampering technology, it is difficult to effectively identify the manipulated data with the single RGB modality. Comparatively, different modalities could be leveraged to collaboratively detect the subtle forgery artifacts. Indeed, there has been a range of work~\cite{wang2022objectformer,qian2020thinking,chen2021local} exploring the integration of multimodal information for media forensics. For tampering detection, several methods reveal that the plain RGB features can be well enhanced by exploiting high-frequency features containing rich manipulation details, as for Deepfake detection, a celebrity video that spread fake news could be easily detected through a comprehensive analysis of visual artifacts and lip-text/lip-audio mis-synchrony. The discovery of more representative modalities and the more effective integration of different modality information are both promising directions for further improvements in forgery detection performance.

\fakeparagraph{Interpretability}: the lack of interpretability of deep neural networks has always been one of their deficiencies despite their powerful discriminative abilities, making it hard for decisions from neural networks to be accepted on serious occasions such as a court. As the current trend of Deepfake and tampering detection methods is still based on deep learning, more efforts need to be made to add more interpretabilities and trustworthiness to these methods. Comparatively, rule-based tampering detection methods, especially the intrinsic pattern analysis, introduced in Sec.~\ref{subsubsec:tampering_rule_detection} have better interpretabilities since they rely on concrete rules rather than data or complex models full of parameters.
As hybrid methods prove feasible (See Sec.~\ref{subsubsec:tampering_hybrid_detection}), the future of a truly reliable forensic tool for tampering detection may constantly revolve around rule-based methods and deep learning.

\section{Conclusion}
\label{sec:conclusion}
This paper presents a comprehensive survey of deep learning-based media forensic methods. Depending on the content of suspicious media data, we divide media manipulation detection into tampering detection and Deepfake detection, which share a wide variety of commonalities in the capture of forgery traces. Public datasets used to benchmark different detection methods and a multitude of forgery detection techniques are presented in detail. Finally, we also discuss current challenges in tampering detection, and provide some insights into future research.

\bibliographystyle{IEEEtran}
\bibliography{main}

\end{document}